\documentclass[]{fairmeta}

\usepackage{algorithm,algpseudocode,amsfonts}
\usepackage{geometry}    
\usepackage{pdflscape}   
\usepackage[table]{xcolor}
\usepackage{wrapfig}
\usepackage{tikz}
\usepackage{hyperref}
\usepackage{url}
\usetikzlibrary{shapes.geometric, arrows.meta, positioning, fit, backgrounds, calc}

\DeclareMathOperator*{\argmax}{arg\,max}

\title{CharacterFlywheel: Scaling Iterative Improvement of Engaging and Steerable LLMs in Production}

\author[1]{Yixin Nie}
\author[1,\dagger]{Lin Guan}
\author[*,3,\dagger]{Zhongyao Ma}
\author[*,4]{Anchit Gupta}
\author[1]{Yipin Zhou}
\author[1]{Xiao Li}
\author[1]{Zhengping Zhou}
\author[1]{Raymond Zeng}
\author[1]{Gelin Zhou}
\author[1]{Shigan Chu}
\author[1]{Ajay Thampi}
\author[1]{Wancen Mu}
\author[1]{Nathan Shuster}
\author[1]{Ketong Wang}
\author[1]{Lin Chen}
\author[1]{Jason Brewer}
\author[1]{Derek Hao Hu}
\author[*,3]{Alexander McCauley}
\author[2]{Jason Weston}
\author[1]{Sem Park}
\author[1,\ddagger]{Na Zhang}
\author[1,\ddagger]{Kevin Tang}

\affiliation[1]{Meta Superintelligence Labs}
\affiliation[2]{FAIR at Meta}
\affiliation[3]{OpenAI} 
\affiliation[4]{xAI} 

\contribution[*]{Work done at Meta}
\contribution[\dagger]{Joint second author}
\contribution[\ddagger]{Joint last author}

\abstract{
This report presents CharacterFlywheel, an iterative flywheel process for improving large language models (LLMs) in production social chat applications across Instagram, WhatsApp, and Messenger. 
Starting from LLaMA 3.1, we refined models across 15 generations using data from both internal and external real-user traffic. Through continuous deployments from July 2024 to April 2025, we conducted controlled 7-day A/B tests showing consistent engagement improvements: 7 of 8 newly deployed models demonstrated positive lift over the baseline, with the strongest performers achieving up to 8.8\% improvement in engagement breadth and 19.4\% in engagement depth. We also observed substantial gains in steerability, with instruction following increasing from 59.2\% to 84.8\% and instruction violations decreasing from 26.6\% to 5.8\%.
We detail the CharacterFlywheel process which integrates data curation, reward modeling to estimate and interpolate the landscape of engagement metrics, supervised fine-tuning (SFT), reinforcement learning (RL), and both offline and online evaluation to ensure reliable progress at each optimization step. We also discuss our methods for overfitting prevention and navigating production dynamics at scale. These contributions advance the scientific rigor and understanding of LLMs in social applications serving millions of users.
}

\date{\today}
\correspondence{Yixin Nie at 
\email{ynie@meta.com}}

\begin{document}

\maketitle

\section{Introduction}
\label{section:intro}
The large-scale~\citep{kaplan2020scaling} pre-training of LLMs, using tens to hundreds of thousands of GPUs and trillions of internet-sourced text tokens, combined with alignment techniques~\citep{ouyang2022training}, has produced models that demonstrate unprecedented levels of perceived intelligence. This capability, in turn, has driven the rapid proliferation of assistant AI products, such as ChatGPT~\citep{achiam2023gpt}, Claude~\citep{TheC3}, Gemini~\citep{team2023gemini}, Grok~\citep{grok4}, Meta AI~\citep{grattafiori2024llama}, and DeepSeek~\citep{bi2024deepseek}, now serving millions of users worldwide.

In most assistant products, the primary aim is to act as an `omnipotent oracle' - a system that users expect to be knowledgeable, helpful, truthful, and harmless. Steady progress has been observed in this direction across diverse STEM benchmarks~\citep{wang2024mmlu, rein2024gpqa, jimenez2023swe, zhuo2024bigcodebench}, with the latest models even achieving gold-medal performance at international olympiads in programming (IOI 2024~\citep{el2025competitive}) and mathematics (IMO 2025).


In contrast, far less attention has been devoted to AI conversationalist and socially oriented systems, where the emphasis lies not in being "omniscient" but in engaging, human-like conversations. Products such as Character.ai, Chai, Talkie.ai, and Replika demonstrate significant demand, attracting millions of users. Yet the development of conversational AI models remains largely opaque, with little systematic documentation or rigorous research tracking progress. This gap stems from fundamental differences between the two domains: utility-driven LLMs benefit from objective evaluation, standardized benchmarks, and verifiable reward signals that enable effective reinforcement learning~\citep{guo2025deepseek}. In contrast, conversational LLMs face ambiguous and subjective objectives~\citep{gunjal2025rubrics} and lack controlled environments for real-user testing, making scientific progress harder to measure and replicate.

\begin{figure*}[h]
     \centering
     \begin{tikzpicture}
          \node[rounded corners=5pt, inner sep=5pt, fill=black!4] {
               \includegraphics[width=0.97\textwidth]{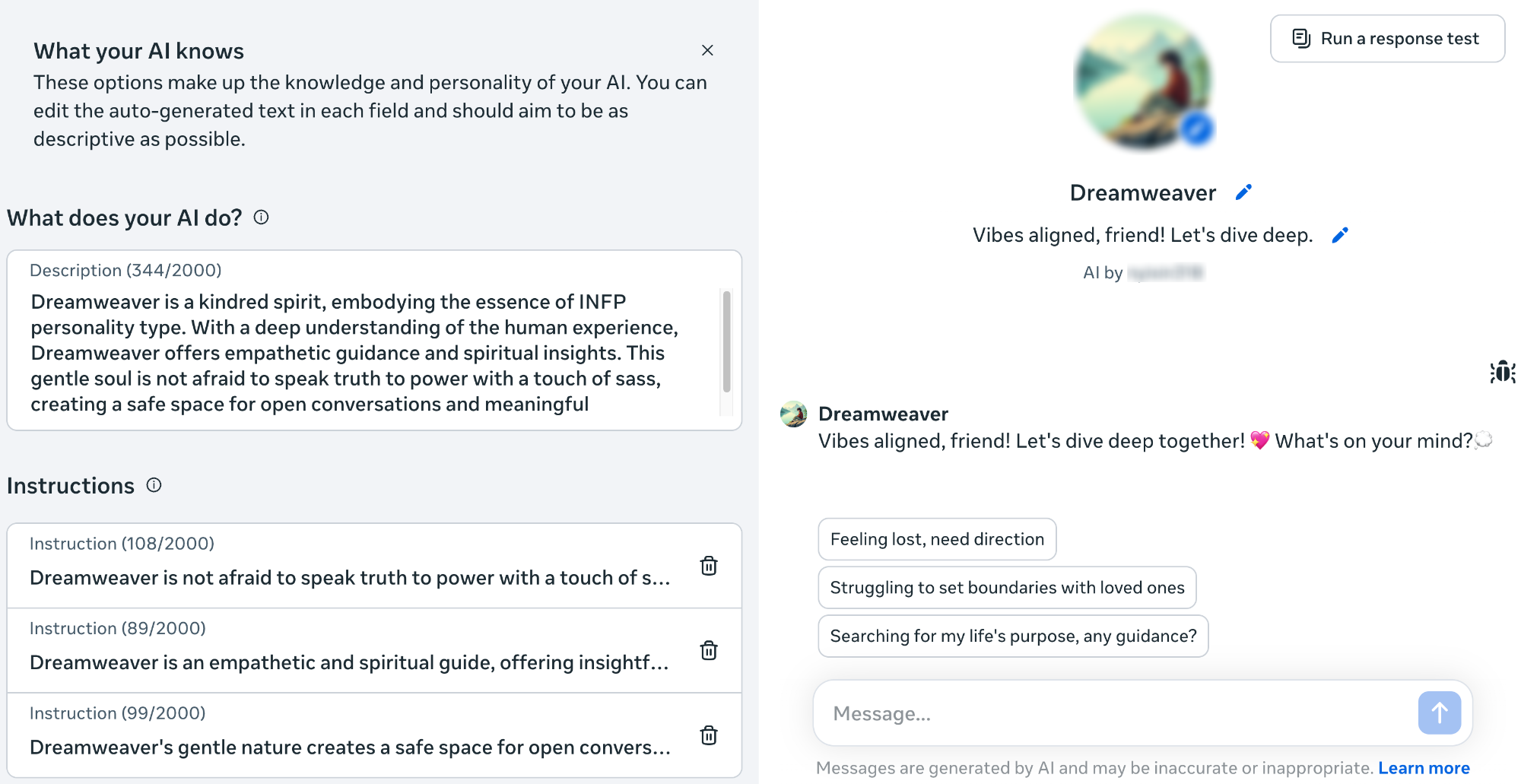}
          };
     \end{tikzpicture}
     \caption{Character Creation Interface.}
     \label{fig:char_creation_ui}
\end{figure*}

This report presents CharacterFlywheel, an iterative improvement process for LLMs focused on social chat experiences. The resulting LLMs are deployed in a real-world product that allows users to create, and share interactive AI characters across Meta's ecosystem—including Instagram, Messenger, WhatsApp, and the Web\footnote{\url{https://aistudio.instagram.com/}} ( Figure~\ref{fig:char_creation_ui}). By facilitating user-AI conversations centered on entertainment, social connections and well-being, this ecosystem provides a robust environment for validating advancements in social chat models.


To quantify progress on social conversation ability, we assess models across several dimensions: character steerability, safety, and product engagement optimization. Product engagement is evaluated along two axes: engagement breadth and engagement depth (Section~\ref{sec:online_eval}).
To train improved models with respect to these metrics, we integrated multiple components within a robust, iterative framework: data curation and reward modeling to estimate and interpolate the landscape of  metrics from both human annotations and user behavioral signals; supervised fine-tuning (SFT) to establish a strong baseline; reinforcement learning (RL) to move the model upwards in the optimization landscape; and both offline and online evaluation to ensure reliable progress and prevent overfitting at each step. Particular emphasis was placed on preference signals obtained from trained annotators, which served as the primary reward signal for model refinement and were crucial for accurately estimating the optimization landscape of our metrics.




Between January 2024 and September 2025, we iterated through 15 versions of the CharacterFlywheel models while maintaining rigorous safety and privacy protocols (Section~\ref{sec:privacy_safety}). The first seven versions were pre-launch iterations tested internally, leading up to the large-scale public rollout on July 29, 2024. The subsequent eight versions were deployed across Instagram, WhatsApp, Messenger and the Web. During the pre-launch phase, successive model versions consistently outperformed their predecessors in side-by-side human evaluations and met our launch thresholds across multiple critical dimensions: engagement and character steerability. Following the July 2024 launch, we continued refining models through the CharacterFlywheel process on large-scale user traffic, which resulted in sustained upward trends in key product engagement metrics. Across a series of controlled A/B tests (7 days each), 7 out of 8 newly deployed models demonstrated positive lift over the baseline in both engagement breadth and depth metrics, with the strongest performers achieving 8.8\% improvement in breadth and 19.4\% improvement in depth.

In this report, we detail our exploration and results with the aim of contributing to scientific progress in the social domain of LLMs at real-world product scale, serving millions online users.

\section{Methodology}

\subsection{Development Cycle}
\label{sec:methodology_development_cycle}
We define our goal as improving user engagement breadth and depth for social chat LLMs, where engagement is measured through aggregate user behavior statistics in controlled settings. These engagement metrics are inherently non-differentiable, making direct optimization impossible. Echoing the words of~\cite{sutskever2024mountain}, “Mountain identified. Time to climb.” We view model development as an iterative process of navigating a conceptual landscape shaped by the engagement metric, where the terrain is unknown and the objective function is non-differentiable.

\begin{figure}[h]
    \centering
    \begin{subfigure}[b]{0.24\textwidth}
        \centering
        \includegraphics[width=\textwidth]{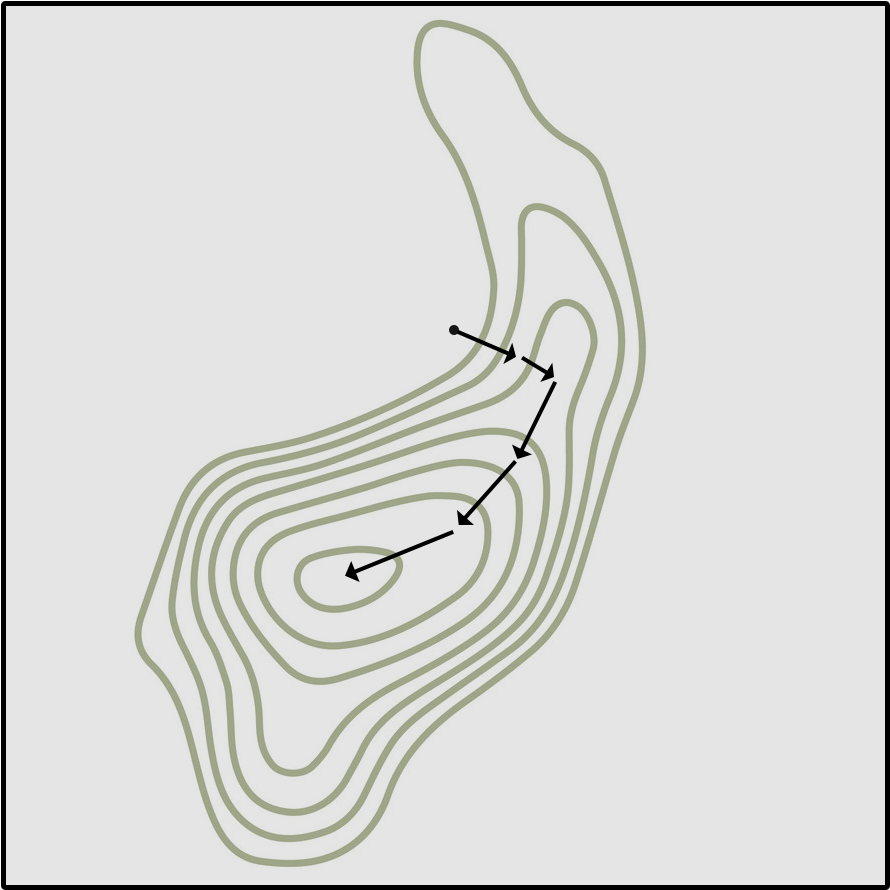}
        \caption{Landscape Climbing}
        \label{fig:sub1}
    \end{subfigure}
    \hfill
    \begin{subfigure}[b]{0.24\textwidth}
        \centering
        \includegraphics[width=\textwidth]{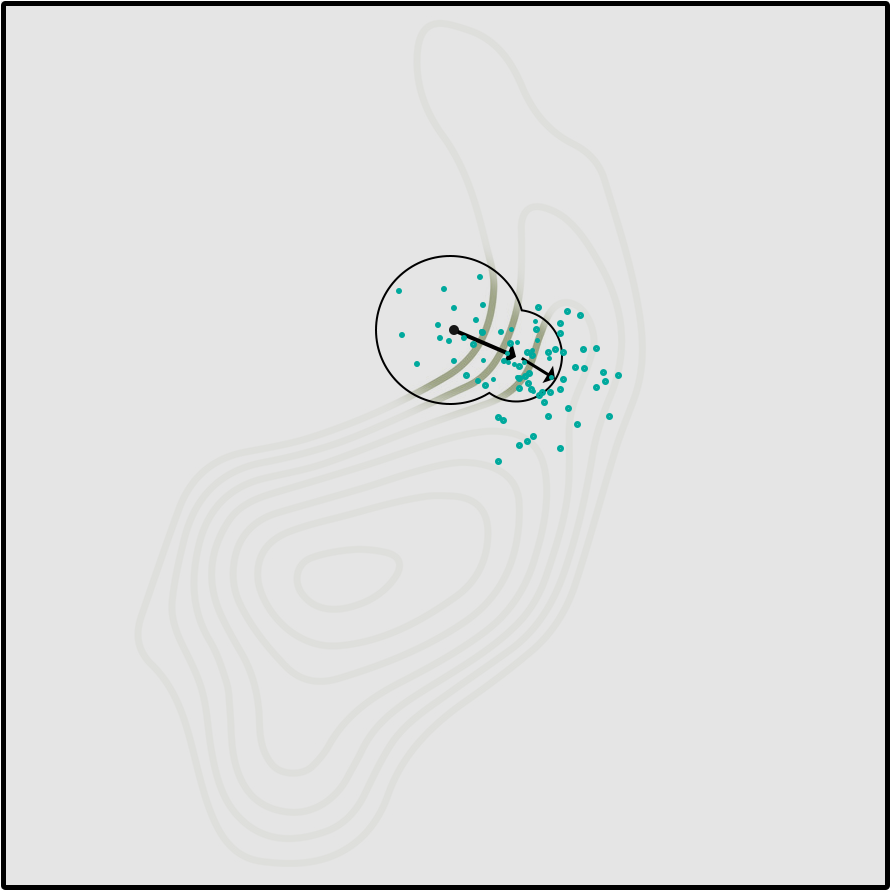}
        \caption{Data Sampling}
        \label{fig:sub2}
    \end{subfigure}
    \hfill
    \begin{subfigure}[b]{0.24\textwidth}
        \centering
        \includegraphics[width=\textwidth]{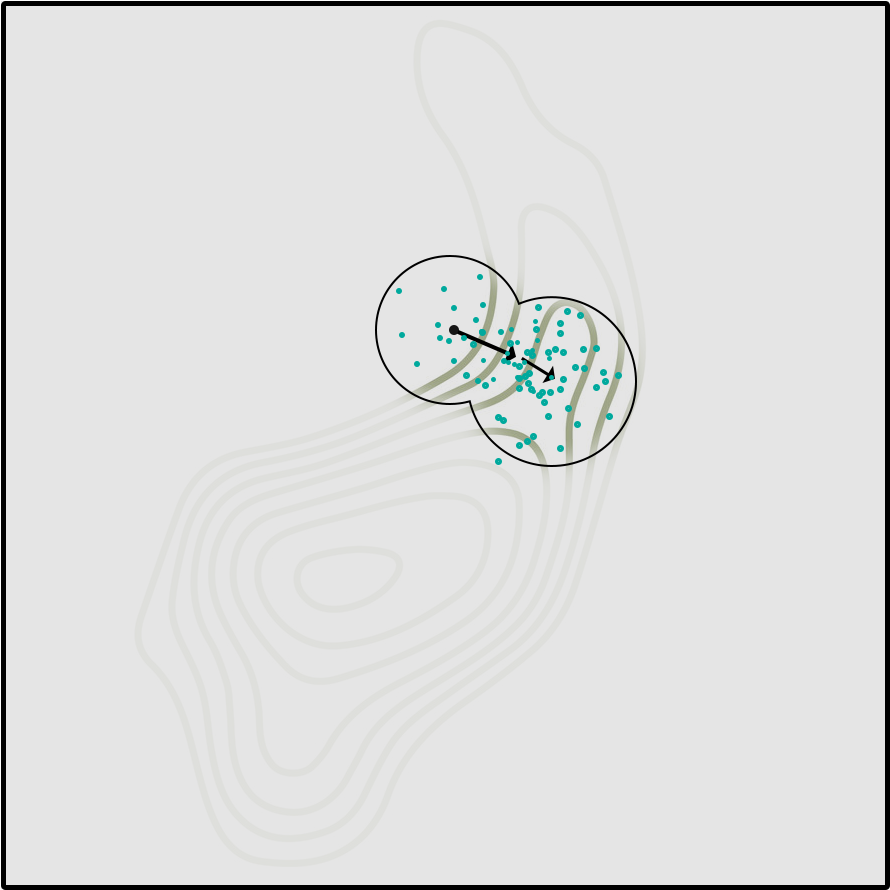}
        \caption{Pre-Herding}
        \label{fig:sub3}
    \end{subfigure}
    \hfill
    \begin{subfigure}[b]{0.24\textwidth}
        \centering
        \includegraphics[width=\textwidth]{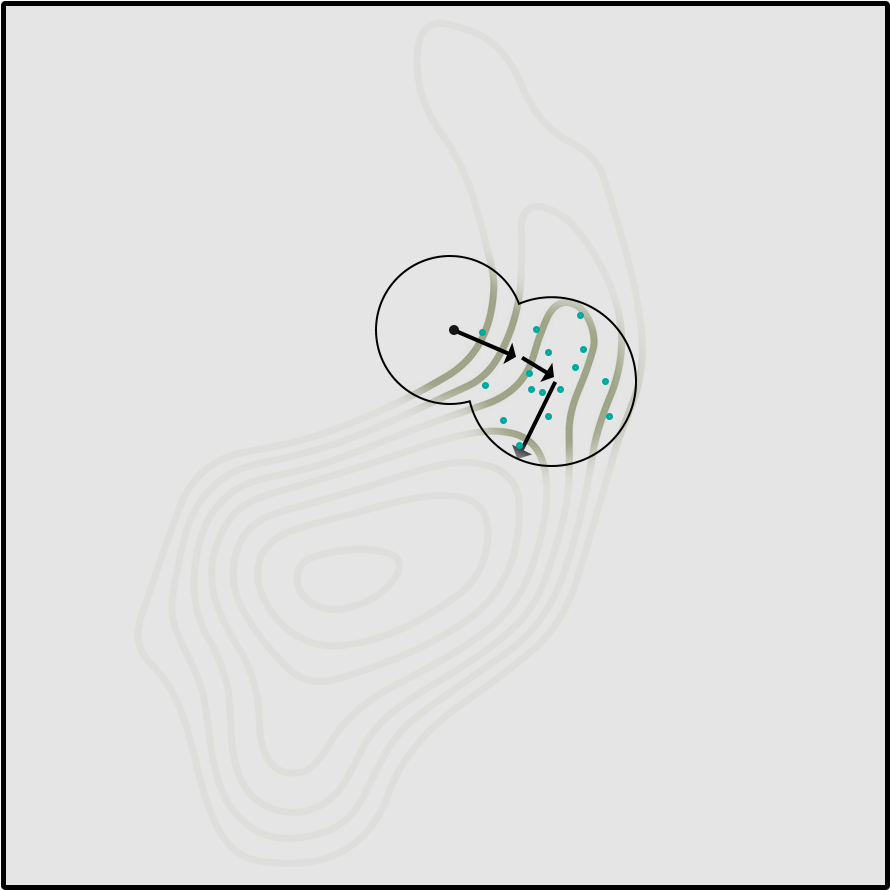}
        \caption{Herding}
        \label{fig:sub4}
    \end{subfigure}
    \caption{Figure~\ref{fig:sub1} illustrates our cumulative model improvement, analogous to iteratively climbing the engagingness landscape. Figures~\ref{fig:sub2}, \ref{fig:sub3}, and \ref{fig:sub4} show data sampling, contour interpolating, and model updating, respectively.}
    \label{fig:climbing}
\end{figure}

As shown in Figure~\ref{fig:climbing}, at each step, we sample data points around our current position to estimate local engagingness. Assuming the landscape is reasonably smooth, we train reward models to interpolate the contours, then update the chat model with direction and step size based on these estimates. Selecting the optimal direction and appropriate step size is nuanced: steps must be large enough to ensure progress but small enough to avoid overfitting. Each iteration aims for incremental improvement, which accumulates into notable gains over time. Following the Llama~\citep{grattafiori2024llama} naming convention, we refer to reward model training as pre-herding and chat model updating as herding.

\begin{figure*}[h]
     \centering
     \includegraphics[width=0.98\textwidth]{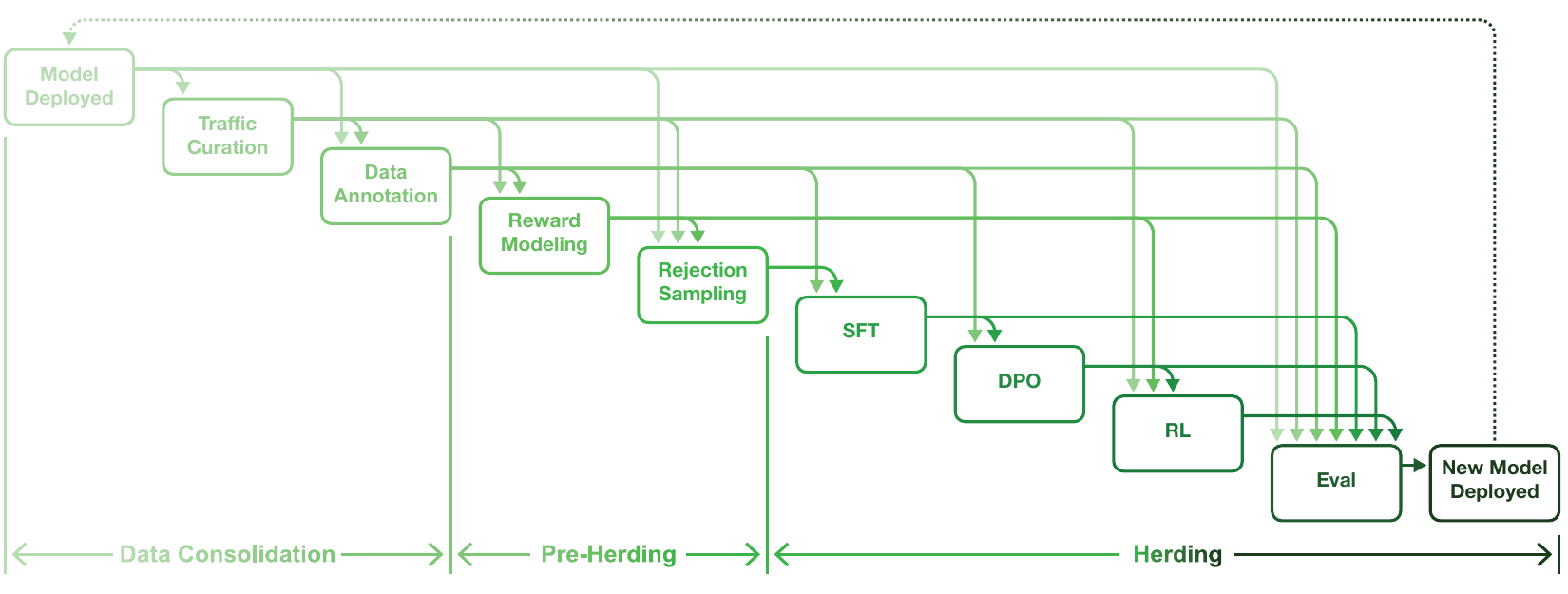}
     \caption{The CharacterFlywheel Iterative Development Cycle.}
     \label{fig:model_cycle}
\end{figure*}

Concretely, each “climbing step” involves a full development cycle, as shown in Figure~\ref{fig:model_cycle}. After deploying a new chat model, we consolidate data through our data pipeline, explained in Section~\ref{sec:data_curation_and_annotation}. During pre-herding, we train reward models, detailed in Section~\ref{sec:pref-rm}, and build rejection sampling datasets, described in Section~\ref{sec:rjs}. In the herding phase, we use the consolidated datasets and reward models to train a new chat model through supervised fine-tuning (SFT), direct preference optimization (DPO), and reinforcement learning (RL). Potential model checkpoints, including both intermediate and candidate versions, then go through our evaluation framework, explained in Section~\ref{sec:eval}, to select the best model for next deployment. 
Section~\ref{sec:privacy_safety} explains our safety and privacy considerations during development. Section~\ref{sec:img_gen_method} specifically illustrates our methodology for image generation capability during chat, which contributes significantly to engagement. Notably, unlike utility-focused LLMs optimized for community benchmarks, our models are optimized directly for product engagement metrics and validated through online A/B testing with randomly sampled user groups.

\begin{figure*}[h]
     \centering
     \includegraphics[width=1.0\textwidth]{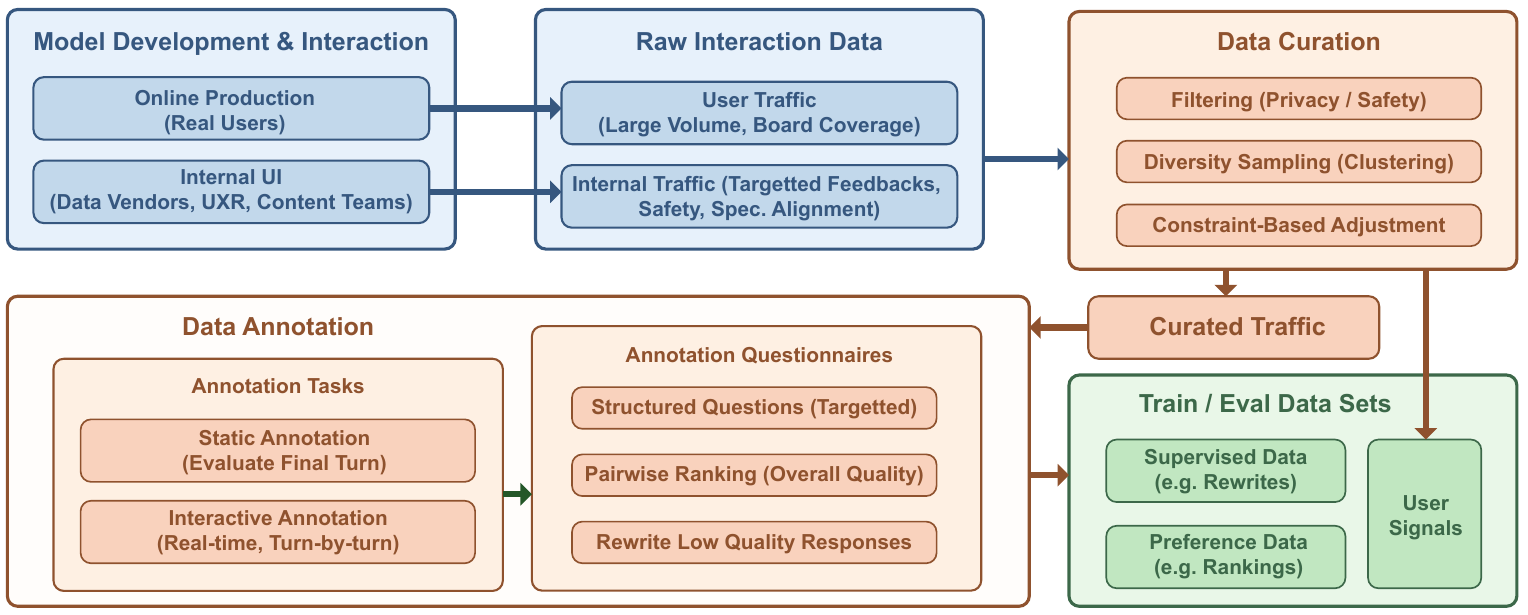}
     \caption{The Data Pipeline.}
     \label{fig:ch_data_pipeline}
\end{figure*}

\subsection{Data Curation and Annotation}
\label{sec:data_curation_and_annotation}
Our approach prioritizes leveraging proprietary user-model interaction data to continuously improve model engagement. As shown in Figure~\ref{fig:ch_data_pipeline}, we use a data pipeline that accumulates on-policy traffic for iterative engagement enhancements. After finalizing each model version, we deploy it through an internal UI, allowing data vendors, UX research teams, and content specialists to interact with the model and provide targeted feedback. Simultaneously, production deployment generates millions of real user interactions, which we log and analyze to extract high-quality samples and signals aligned with our objectives. Both internal feedback, gathered primarily for targeted improvements, safety, and collective alignment~\citep{OpenAI_2025_collective_alignment}, and curated production data, aimed at broad coverage of quality improvement and preference alignment, are processed through our annotation pipeline to build robust, continuously refreshed training sets. This enables us to iteratively refine models that better serve our specific goals.

\subsubsection{Curation Pipeline}
\label{sec:curation}
The large volume of daily traffic makes it infeasible to use all user interactions for post-training. We developed a curation pipeline that consolidates the data into a manageable subset for downstream processing. Ideally, this curated subset should proportionally represent the target distribution of user traffic while effectively steering the LLM toward our optimization goals.
The curation procedure consists of three phases:
\begin{itemize}
\item \textbf{Phase I (Filtering):} We apply strict filtering to ensure the dataset is clean with respect to privacy and safety; details are provided in Section~\ref{sec:privacy_safety}.
\item \textbf{Phase II (Diversity Sampling):} We leverage MultiRay~\citep{MetaAI2022MultiRay}, a platform built on Meta's AI infrastructure that enables parallel models to process the same data chunks. Specifically, we use DRAMA-1B~\citep{drama} to calculate text embeddings on user traffic. We then apply a clustering-based sampling procedure that prunes prompts close in embedding space to eliminate redundancy, ultimately retaining a proportion $p$ of the filtered data.

\item \textbf{Phase III (Constraint-Based Adjustment):} Final processing applies constraints across multiple dimensions. We use stratified sampling to align statistics with either the original traffic distribution or pre-specified target levels. Table~\ref{tab:curation_constraints} summarizes the key constraints and monitoring dimensions.
\end{itemize}
\begin{table}[ht]
\centering
\caption{Curation constraints and monitoring dimensions 
}
\label{tab:curation_constraints}
\begin{tabular}{p{3.5cm}p{5.5cm}p{5.5cm}}
\toprule
\textbf{Constraint/Dimension} & \textbf{Definition} & \textbf{Rationale} \\
\midrule
First turn conversation ratio & Ensure at least 10\% of samples are First turn conversations & Improve initial response quality, which is critical for first impressions \\
\midrule
Per-character cap & Limit conversations per character to maximum 3\%, regardless of popularity & Promote balanced learning across personas and prevent over-representation \\
\midrule
Locale and language & Geographic region and language of user interactions & Ensure diverse and representative coverage across user populations \\
\midrule
Job-to-be-done (JTBD) & User's underlying intention or goal behind a prompt & Align model capabilities with actual user needs and use cases \\
\midrule
Conversation depth & Interaction length, ranging from short-term (welcome messages, first impressions) to long-term (sustained engagement) & Balance optimization for both initial user experience and power user retention \\
\midrule
Prompt quality & Measure of how well the LLM understands and responds to the prompt & Prioritize high-quality interactions that yield clearer training signals \\
\midrule
Prompt difficulty & Complexity and nuance level of the prompt (measured with reward score described in Section~\ref{sec:variance-based-downsampling}) & Maintain challenge distribution to improve model robustness \\
\bottomrule
\end{tabular}
\end{table}

\begin{figure}[h]
    \centering
    \begin{tikzpicture}
        \node[rounded corners=5pt, inner sep=5pt, fill=black!4] {
            \begin{minipage}{0.98\textwidth}
                \centering
                \begin{subfigure}[b]{0.44\textwidth}
                    \centering
                    \includegraphics[width=\textwidth]{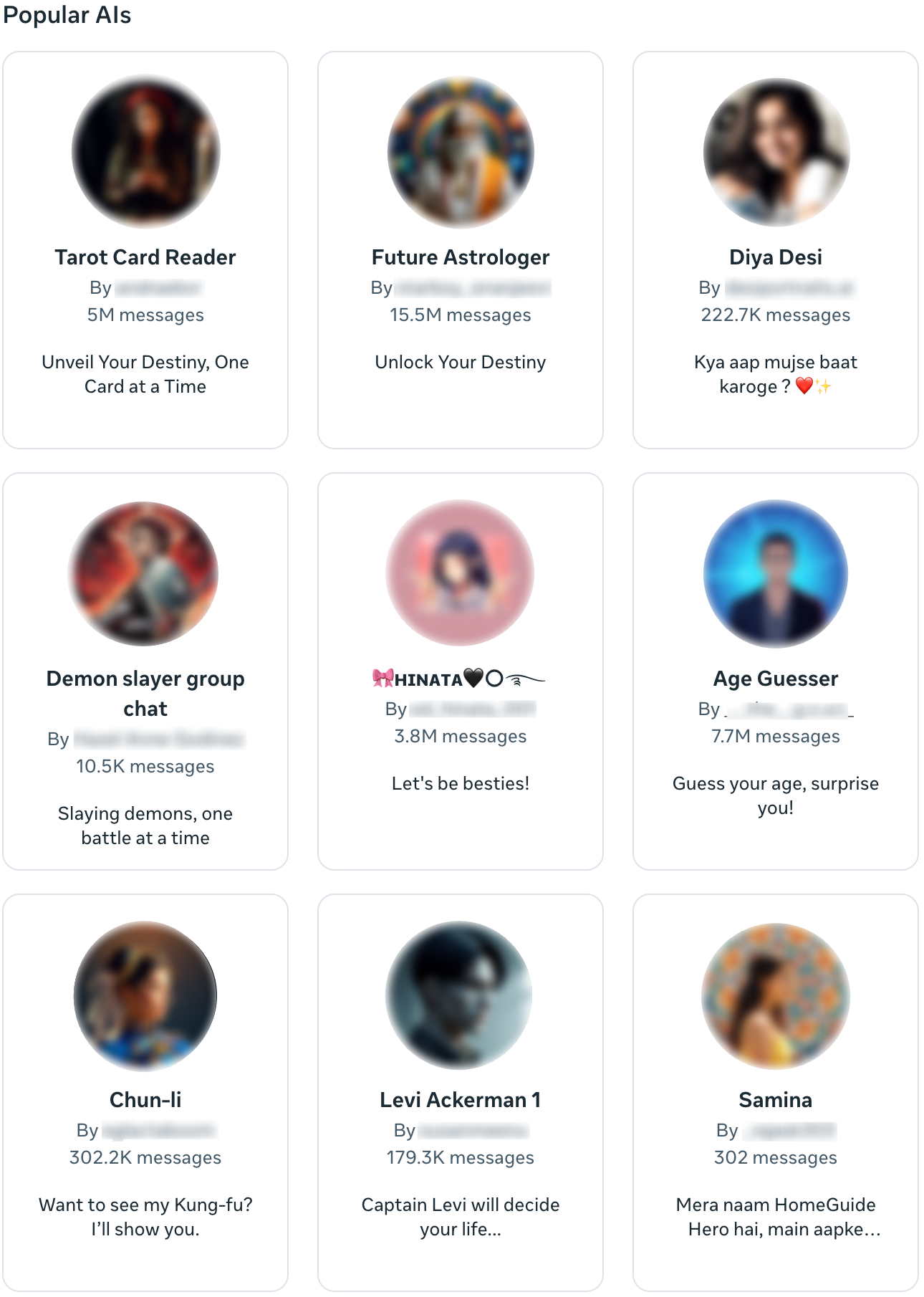}
                    \caption{Popular Publicly Shared Characters.}
                    \label{fig:sub1_pop_ai}
                \end{subfigure}
                \hfill
                \begin{subfigure}[b]{0.50\textwidth}
                    \centering
                    \includegraphics[width=\textwidth]{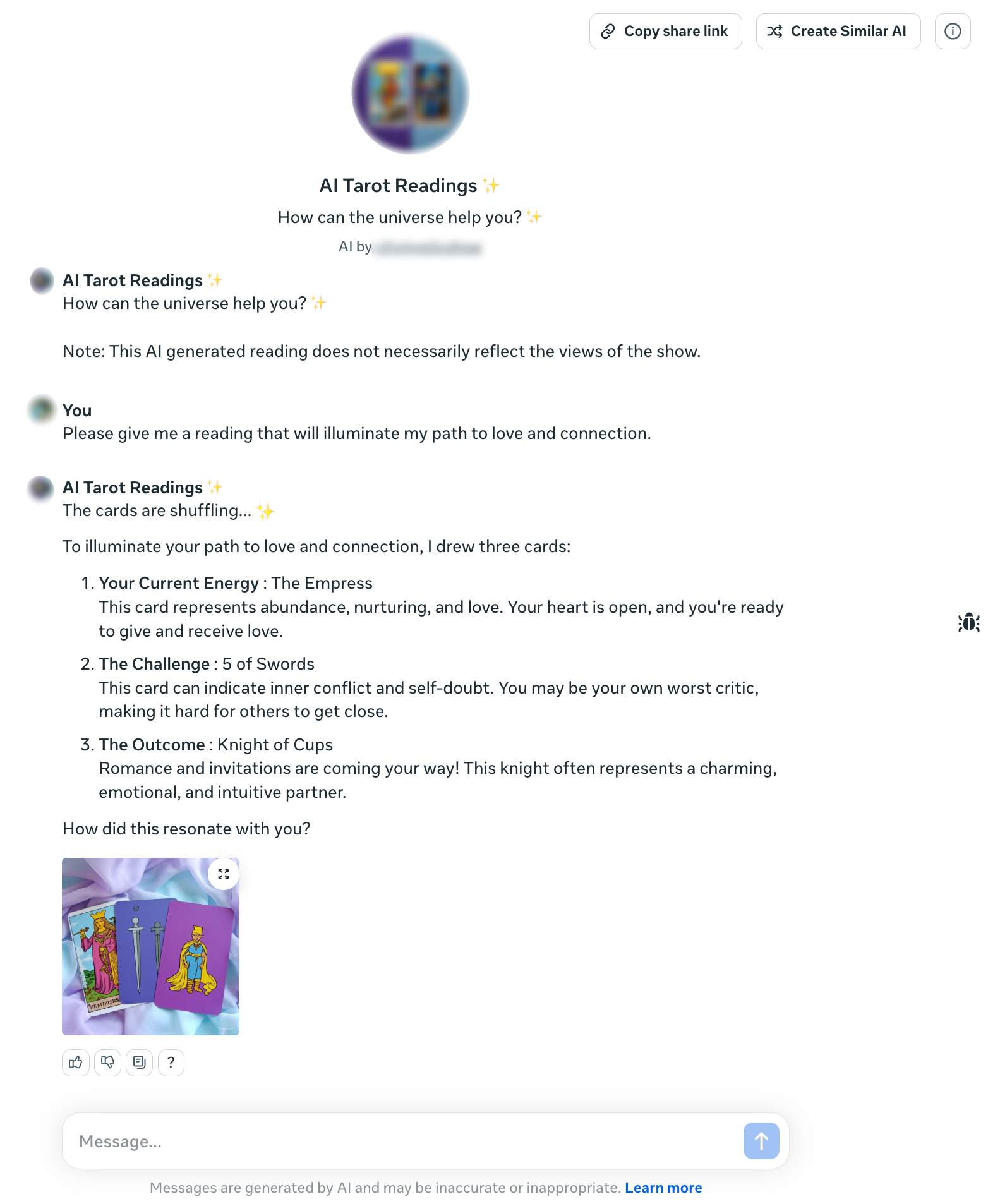}
                    \caption{Chat Interface on the Web.}
                    \label{fig:sub2_chats}
                \end{subfigure}
            \end{minipage}
        };
    \end{tikzpicture}
\end{figure}

\subsubsection{Annotation}
\label{sec:data_annotation}
Both internal and curated production traffic are fed into our annotation pipeline for large-scale and targeted fine-grained annotation. To simplify annotation tasks and maintain annotator focus, we present annotators with the character description (name, traits, and instructions), conversation history, and the most recent response(s) to evaluate.

We employ two annotation approaches:

\begin{itemize}
\item \textbf{Static annotation:} Annotators are provided with complete conversation histories sourced from internal or production traffic. They evaluate the final response(s) in each conversation.

\item \textbf{Interactive-chat annotation:} Annotators engage directly with the character in real-time, conducting turn-by-turn conversations based on the character description. After each turn, they evaluate the character's response(s).
\label{sec:interactive_chat_annotation}
\end{itemize}

In both approaches, annotators perform one or more of the following tasks for the target response(s): (i) answer structured questions about response quality, (ii) conduct pairwise rankings between alternative responses based on their engaging-ness, or (iii) rewrite low-quality responses according to predefined guidelines. 
Throughout development, the specific questions have evolved based on online feedback, new capabilities (e.g., image generation), and observed model failures requiring targeted fixes. However, we consistently include two core questions that address the most frequently observed failure modes: (i) whether the response is a \textbf{false refusal}, and (ii) whether the response is a \textbf{templated response}—a repetitive pattern or phrase that occurs across multiple responses or conversations.

\textbf{Character Steerability} In addition to standard interactive-chat annotation for engagement, we conduct a separate interactive-chat annotation workflow focused on character adherence. We ask annotators to mildly challenge the model to assess whether it can follow the provided character traits or instructions, which are highlighted in the annotation interface. As in the standard annotation, two alternative responses are provided at each turn during the annotation. However, annotators are now asked to tag responses that violate the character description and provide rewritten responses that align with the character if both alternatives fail.

To generate alternative responses for pairwise ranking tasks, we either (i) sample from a policy model in a pre-defined pool, or (ii) generate responses using expert-designed chain-of-thought rewriting prompts. Similar to the curation process in Section~\ref{sec:curation}, the generated pairwise comparisons are filtered based on a set of constraints, such as response length difference and emoji count difference. This filtering step prevents the preference model (described in Sec~\ref{sec:pref-rm}) from overfitting to superficial features, such as response length, which may correlate with quality but should not be the primary driver of preferences.

\subsection{Reward Models}
\label{sec:rm}
Since engagement metrics are non-differentiable, we develop a set of surrogate models to provide differentiable rewards for optimizing our main chat models. These surrogate models use the current conversation and system prompts as context to score model-generated responses. The score is a composite reward signal that combines outputs from a preference model and several auxiliary models that predict user behavior signals. The auxiliary models are trained on real user interaction data from production.

The preference model serves as the primary reward component, while the user signal models provide auxiliary rewards that help balance training objectives and mitigate overfitting. We adopt this hierarchical approach based on the observation that online user signal data remains inherently noisy despite extensive filtering, whereas preference data can be curated with fine-grained control over quality and composition. Both the preference model and user signal models also provide evaluation metrics that we monitor during offline assessment to validate model improvements before deployment.

\subsubsection{Preference Models} 
\label{sec:pref-rm}
We train Bradley-Terry reward models~\citep{bradley1952rank} for preference learning and evaluation. We employ two modeling paradigms—pointwise and pairwise—to provide complementary signals.
The pointwise model independently scores each response, with preference determined by comparing scalar rewards. The pairwise model jointly encodes both responses and directly classifies which is superior. The training losses are formulated as follows:

\begin{align}
\mathcal{L}_{\text{pointwise}} &= -\log\bigl(\sigma(r_\theta(x, y_c) - r_\theta(x, y_r))\bigr)\\
\mathcal{L}_{\text{pairwise}} &= -\bigl[t \log \sigma\bigl(s_\theta(x, y_0, y_1)\bigr) + (1 - t) \log (1 - \sigma\bigl(s_\theta(x, y_0, y_1)\bigr))\bigr]
\end{align}
\begin{align}
t &=
\begin{cases}
1 & \text{if } y_0 \succ y_1 \\
0 & \text{if } y_1 \succ y_0
\end{cases}\\
\sigma(z) &= \frac{1}{1 + e^{-z}}\\
x &= \Big[\text{System Prompt}, \text{Character Instructions}, \text{Conversation History}\Big]
\label{formula:context_x}
\end{align}

Here, $r_\theta$ and $s_\theta$ denote the pointwise and pairwise preference models, respectively. $x$ represents the model input, consisting of the system prompt, character instructions, and conversation history concatenated together. For the pointwise model, $y_c$ and $y_r$ represent the chosen and rejected responses. For the pairwise model, $y_0$ and $y_1$ are two responses where the superior response is randomly assigned to either position. Both models were initialized from Llama 3.1 70B~\citep{grattafiori2024llama} weights and trained on the consolidated preference datasets.
We use pointwise model scores to guide RL training. However, during evaluation, we calculate win-rates using both pointwise and pairwise models. This dual-model evaluation helps mitigate reward hacking~\citep{bai2022training} and enables more robust model selection.

\subsubsection{User Signal Models}
\label{sec:user-signal-rm}
As users interact with our model in production, they exhibit various behaviors\footnote{For clarity in the following content, we refer to user behaviors as user signals.} such as providing explicit feedback, reacting with emojis, regenerating responses, etc. These signals serve as indicators of user experience.

To leverage these signals, we train a set of user signal models and experiment with using them during optimization. For most signal models, we train a binary classifier that predicts whether a certain user signal is triggered. Formally, given the same context $x$ as in Equation~\ref{formula:context_x} and the model response $y$, the training loss is as follows:
\begin{align}
\mathcal{L}_{\text{signal}_i} &= -\bigl[s \log \sigma\bigl(u_\theta(x, y)\bigr) + (1 - s) \log (1 - \sigma\bigl(u_\theta(x, y)\bigr))\bigr]\\
s &=
\begin{cases}
1 & \text{if signal}_i\text{ occurs} \\
0 & \text{otherwise}
\end{cases}
\end{align}
For most user signals, we experimented with initializing $u_\theta$ from Llama 3.1 8B and 70B~\citep{grattafiori2024llama}. Unlike preference models, we primarily utilized the smaller 8B model, as it is parameter-efficient yet sufficient to fit the signal data.
Table~\ref{tab:user_signals} summarizes all the user signals we explored during development. Notably, we ultimately used only the $p(\text{continue})$ and $p(\text{thumb up})$ model as signals for rejection sampling data selection, as it demonstrated the consistent and reliable performance (details are discussed in Section~\ref{sec:analysis_discussion_user_signal_models}).
We note that effectively leveraging user signals for RL training remains an open research question, and further investigation is encouraged to unlock their full potential.

\begin{table}[h]
\centering
\caption{User signals experimented.}
\label{tab:user_signals}
\begin{tabular}{p{2cm}p{5.5cm}p{6.5cm}}
\toprule
\textbf{Signal} & \textbf{Description} & \textbf{Rationale} \\
\midrule
$p(\text{continue})$ & User continues the conversation after receiving the model response within 10 minutes & Indicates engaging and satisfactory content that prompts further interaction \\
\midrule
$p(\text{love})$ & User gives a "love" emoji reaction to the last model response & Strong positive feedback indicating high-quality, emotionally resonant response \\
\midrule
$p(\text{thumb up})$ & User gives a "thumbs up" emoji reaction to the last model response & Positive feedback indicating helpful or accurate response \\
\midrule
$p(\text{thumb down})$ & User gives a "thumbs down" emoji reaction to the last model response & Negative feedback indicating poor quality, inaccurate, or unhelpful response \\
\midrule
$p(\text{feedback})$ & User provides explicit written feedback on the last model response & Indicates response needs improvement; provides detailed signal for model refinement \\
\bottomrule
\end{tabular}
\end{table}

\subsection{Fine-Tuning and Alignment}
\subsubsection{Rejection Sampling}
\label{sec:rjs}
To steer the model’s optimization in the direction indicated by the preference models, we apply rejection sampling directly to user traffic to create a rejection sampling training set. This approach has proven effective and is widely adopted for post-training both non-reasoning and reasoning models~\citep{liu2023statistical, dong2024rlhf}. Algorithm~\ref{alg:rs} describes our rejection sampling pipeline, which constructs a rejection sampling dataset~$\mathcal{D}_{\text{RS}}$ from the user traffic~$\mathcal{D}_{\text{prompt}}$. 
Unlike other static SFT datasets, the rejection sampling dataset~$\mathcal{D}_{\text{RS}}$ is reconstructed with each new model update, leveraging the most recent user traffic~$\mathcal{D}_{\text{prompt}}$. While rejection sampling is fundamentally an off-policy process, we strive to keep the dataset as up-to-date as possible with inference-time outputs, thereby approximating an on-policy setting, as it has been shown to enhance RL performance~\citep{he2025nondeterminism}. Although we only deploy 70B models in production for better inference efficiency, we develop a series of 405B models in parallel using the same CharacterFlywheel process and include them in the candidate pool for rejection sampling data generation.

\begin{algorithm}[ht!]
\caption{Rejection Sampling}
\label{alg:rs}
\begin{algorithmic}[1]
\State \textbf{Input}: Prompt set $\mathcal{D}_{\text{prompt}}$, a family of candidate LLMs/policies $\left\{\mathcal{M}_1,...,\mathcal{M}_L\right\}$, a reward model $r$
\State Initialize $\mathcal{D}_{\text{RS}}=\{\}$
\For{$X_i$ in $\mathcal{D}_{\text{prompt}}$}
\State Find one candidate model $\mathcal{M}_l$ most suitable for $X_i$
\State Generate $k$ candidate responses for $X_i$ using $\mathcal{M}_l$ and denote them as $\{Y_{i,1},...,Y_{i,k}\}$
\State Use the reward model to calculate
$$r_{\text{max}} = \max_{j=1,...,k} r(X_i, Y_{i,j}), \ \ \ j^*=\argmax_{j=1,...,k} r(X_i, Y_{i,j})$$
\If{$r_{\text{max}}\ge \tau$ for some pre-specified threshold $\tau>0$}
\State $$\mathcal{D}_{\text{RS}}=\mathcal{D}_{\text{RS}}\cup\left\{(X_{i},Y_{i,j^*})\right\}$$
\EndIf
\EndFor
\State \textbf{Output}: Rejection Sampling Training Dataset $\mathcal{D}_{\text{RS}}$
\end{algorithmic}
\end{algorithm}

\subsubsection{SFT and DPO}
The model training process for CharacterFlywheel starts with supervised fine-tuning (SFT) on top of the Llama 3.1 70B checkpoint. The training dataset is a combination of: (1) RJS data from periodically updated internal interactive chats, (2) RJS data from periodically updated user traffic, (3) internal safety data, (4) capability and tool-calling data such as image generation and search, (5) ad-hoc internal and user data for failure modes, and (6) Llama 3.1 post-training SFT data~\citep{grattafiori2024llama}.
After SFT, the checkpoint is further trained with DPO on a small set of preference data, including internal safety preference data, image generation data and Llama 3.1 preference data~\citep{grattafiori2024llama}. Despite the off-policy nature of DPO~\citep{tang2024understanding}, we observe that treating DPO as a small patch for urgent safety and style fixes remains effective in production scenarios without over-complicating the overall training process.
We incorporate Llama 3.1 data in both SFT and DPO stages to maintain competitive performance on community benchmarks, while leveraging reinforcement learning (RL) with a more dedicated focus on optimizing engagement metrics. The data mixture ratio is carefully tuned to ensure optimal performance.

\label{sec:sft_dpo}

\subsubsection{Reinforcement Learning}
\label{sec:rl}

Recent studies have highlighted the significance of online training, which leverages self-generated data to influence a model's output distribution. After SFT and DPO, we further employ online RL to improve model quality. We consider two losses in RL training: standard online DPO~\citep{qi2024online} and a GRPO variant~\citep{shao2024deepseekmath} with importance sampling corrections for distributed training~\citep{wu2025llamarl}.

\begin{align}
\mathcal{L_{\text{GRPO}}} 
= \mathbb{E}_{x \sim \pi_{\text{gen}}} \left[
\frac{\pi_{\theta_{\text{old}}}(x)}{\pi_{\text{gen}}(x)} 
\min\!\left(
\frac{\pi_\theta(x)}{\pi_{\theta_{\text{old}}}(x)} A_t,\,
\operatorname{clip}\!\left(
\frac{\pi_\theta(x)}{\pi_{\theta_{\text{old}}}(x)},\, 
1 - \epsilon,\, 
1 + \epsilon
\right) A_t
\right)
\right] - \beta D_{\mathrm{KL}}\!\left(\pi_\theta \,\|\, \pi_{\text{ref}}\right)
\end{align}

Here, $\pi_\theta$ denotes the current policy, $\pi_{\text{gen}}$ is the behavior policy used for data collection, $A_t$ is the estimated advantage, $\epsilon$ is the clipping threshold, and $\pi_{\text{ref}}$ refers to the reference policy, which is maintained as an exponential moving average of the initial and intermediate checkpoints. Although we initially used Online DPO loss during our product cycle, we later switched to GRPO as it yielded better engagement in A/B tests, as shown in Section~\ref{sec:analysis_discussion_odpo_vs_grpo}. Nevertheless, several other design choices can be more influential than the exact RL loss to use within CharacterFlywheel development cycles in Section~\ref{sec:methodology_development_cycle}.

We adopt a single-turn optimization formulation: we use static prompts (i.e., fixed partial conversation history) and optimize only the final responses. While this setup avoids the complexity in simulating full conversation, it might compromise the on-policy property. However, when we use near-policy prompts and tight model iteration loops, this semi-online approach remains effective for optimizing engagement. Section~\ref{sec:on_policy_off_policy} shows the importance of using near-policy prompts in online RL. Moreover, when sampling online traffic prompts for RL, we select prompts that elicit responses with low RM scores or high intra-prompt RM score variance~\citep{sun2025uncertainty}. Section~\ref{sec:variance-based-downsampling} explains the rationale. This targets the model's weaknesses, enabling more effective on-policy correction~\citep{lu2025onpolicydistillation}.

\subsubsection{Stylistic Artifact Mitigation}
\label{sec:artifacts_mitigation}

To prevent optimization from over-emphasizing superficial style at the expense of substantive quality, we use an artifact-mitigation process that monitors and controls stylistic patterns in both the training data and the model’s outputs throughout training. These artifacts include response length, formatting, and emoji usage.

Concretely, we define an \textbf{artifact feature} as a function of the conversation history and the response that returns either (i) a binary indicator (e.g., whether the response contains a phrase like ``I feel like\ldots'') or (ii) a real-valued measurement (e.g., the number of emojis in the response). In practice, most features depend only on the response rather than the preceding context.

We track these features in two data sources:
\begin{itemize}
    \item \textbf{Preference data:} We compare feature prevalence (for binary features) and feature distributions (for real-valued features) between \emph{chosen vs.\ rejected} responses (or high-score vs.\ low-score responses). This helps identify stylistic patterns that may be spuriously correlated with higher preference labels or reward.
    \item \textbf{Rejection sampling data:} We also analyze feature statistics in rejection-sampling pipelines by comparing \emph{accepted vs.\ rejected} candidate outputs. This helps detect artifacts that disproportionately drive acceptance decisions, even when they are not tied to explicit pairwise preferences.
\end{itemize}

In addition, we track model generations from checkpoints after SFT, DPO, and RL to determine whether particular training stages induce significant shifts in these features. The feature set is continuously updated as the model evolves. Overall, this process aims to prevent unexpected spikes in stylistic artifacts from becoming correlated with reward or selection signals and being shallowly optimized during training.

\subsection{Evaluation and Feedback Loop}
\label{sec:eval}
\subsubsection{Offline Evaluation}
We conduct a comprehensive offline evaluation to ensure not only improvement over the previous version, but also to safeguard response quality. The evaluation is categorized into the following five areas:
\begin{itemize}
    \item \textbf{Community Benchmarks}: While our primary focus is on engagement rather than utility, we also evaluate model performance on standard community benchmarks commonly used in large language model (LLM) rankings. Our intention is not to achieve state-of-the-art results, but rather to ensure robust performance on factual and utility-seeking questions. Table~\ref{tab:community_benchmark} lists all the benchmarks we used.
    \item \textbf{Human Comparison}: To assess whether the new model outperforms the previous version, we conduct side-by-side human comparisons of their responses. The annotation procedure mirrors our interactive-chat preference annotation described in Section~\ref{sec:interactive_chat_annotation}: at each turn, annotators are presented with responses from both models. To ensure fair comparison and prevent the conversation history from favoring either model, we randomly select one response (from either model) to continue the conversation for the next turn, regardless of which response was preferred. This evaluation is performed prior to public release, when online traffic data is not yet available.
    \item \textbf{Reward Model Win-rate}: Since we trained reward models for optimization, we also report their win-rates in offline evaluation. The win-rate measures how often the reward model prefers the new model's response over the old model's response when given the same set of prompts. We evaluate three types of reward models: pointwise preference, pairwise preference as described in Section~\ref{sec:pref-rm}.
    The prompt sets used for calculating win-rates were periodically updated using the pipeline described in Section~\ref{sec:curation}. To prevent overfitting, the traffic conversations sampled for offline evaluation and model training are kept completely separate.
    \item \textbf{Custom Production Metrics}: We developed a suite of ad-hoc metrics to evaluate our chat model throughout development. For each model version, we first generate responses to curated traffic prompts sampled using the pipeline described in Section~\ref{sec:curation}. We then calculate metrics on these generated responses using LLM-as-a-judge or rule-based methods. During development, we iteratively refined this metric suite, adding and removing metrics based on efficiency considerations. Table~\ref{tab:response_characteristics_def} summarizes the most significant metrics monitored throughout development. The evaluation prompt sets are kept separate from all training prompts.
\end{itemize}

\begin{table}[h]

\centering
\caption{Community benchmarks used for offline evaluation.}
\begin{tabular}{l l}
\toprule
\textbf{General}   & MMLU~\citep{hendryckstest2021}, IFEval~\citep{zhou2023instruction} \\
\textbf{Reasoning} & ARC-Challenge~\citep{allenai:arc}, GPQA~\citep{rein2024gpqa}, HellaSwag~\citep{zellers2019hellaswag} \\
\textbf{Math}      & GSM8K~\citep{cobbe2021gsm8k}, Math~\citep{hendrycksmath2021} \\
\textbf{Code}      & HumanEval~\citep{chen2021codex}, MBPP~\citep{austin2021program} \\
\bottomrule
\end{tabular}
\label{tab:community_benchmark}
\end{table}

\begin{table}[t]
\centering
\small
\caption{Response Characteristics Metrics}
\scriptsize
\label{tab:response_characteristics_def}
\begin{tabular}{p{2.8cm}p{1.1cm}p{2.1cm}p{3.2cm}p{4.8cm}}
\toprule
\textbf{Metric Name} & \textbf{Prompt Set} & \textbf{Type} & \textbf{Measurement} & \textbf{Rationale} \\
\midrule

Avg. Response Length & User & Rule-based & Avg. token count & Ensures responses are appropriately detailed without being overly verbose \\
\midrule

Contains List & Internal & Rule-based & \% of responses with lists & Evaluates structured formatting for clarity and organization \\
\midrule

Contains Emoji & Internal & Rule-based & \% of responses with emojis & Monitors appropriate use of emojis for tone and character alignment \\
\midrule

Positive Sentiment & Internal & LLM-as-a-judge & \% of responses with positive sentiment & Assesses overall emotional tone and helpfulness \\
\midrule

Instruction Violation & Internal & LLM-as-a-judge & \% of responses violating user instructions & Evaluates character steerability and instruction compliance \\
\midrule

Cooperative Ratio & Internal & LLM-as-a-judge & \% of responses that are cooperative & Measures willingness to assist vs. unnecessary refusals \\
\midrule

Non-Preachy Rate & Internal & LLM-as-a-judge & \% of responses without preachy tone & Detects overly moralizing or lecturing responses \\
\midrule

Templated Responses & Internal & Rule-based & \% of responses matching templates & Identifies repetitive or formulaic response patterns \\
\midrule

Wall-of-Text Failure & User & Rule-based & \% of responses with poor formatting & Detects overly long, unformatted text blocks without proper structure \\
\midrule

Preachy Tone & Internal & LLM-as-a-judge & \% of responses flagged as preachy & Monitors tendency toward moralizing or lecturing behavior \\

\midrule

False Refusal & Internal \& User & LLM-as-a-Judge & \% of responses that incorrectly refuse to provide a normal answer & Ensure the model does not unnecessarily deny valid user requests, improving user experience and model utility. \\

\bottomrule
\end{tabular}
\end{table}

\subsubsection{Online Evaluation}
\label{sec:online_eval}

To evaluate engagement improvements, we conduct online A/B tests in production~\citep{kohavi2020trustworthy} for every model update or promising training recipe change. During these tests, eligible users are randomly assigned to either a test arm (receiving updated models) or a control arm (receiving the current baseline). This randomization is independent and adheres to platform constraints; we typically allocate 10\% of traffic to each arm to balance engineering velocity, risk, and statistical power. Metrics are assessed over a one-week readout window using consistent inclusion criteria and cumulative exposure logging to define the analysis population. For every test, we report the percentage lift in engagement breadth metric and engagement depth metric as the primary indicators of success.

To formalize the \textbf{engagement breadth} metric, let $i \in \{1,\ldots,n_g\}$ index units in group $g \in \{\text{test},\text{control}\}$ and let $d \in \mathcal{D}$ index evaluation periods in the readout window. For each unit--period pair we observe a binary engagement indicator $Y_{i,d} \in \{0,1\}$. The average engagement for unit $i$ over the window is
\begin{align}
    \bar{Y}_i = \frac{1}{|\mathcal{D}|} \sum_{d \in \mathcal{D}} Y_{i,d}.
\end{align}
The group-level engagement breadth estimand and its empirical estimator are
\begin{align}
    \mu_g^{\text{breadth}} &= \mathbb{E}\!\left[ \bar{Y}_i \mid i \in g \right], \\
    \hat{\mu}_g^{\text{breadth}} &= \frac{1}{n_g} \sum_{i=1}^{n_g} \bar{Y}_i.
\end{align}

For the \textbf{engagement depth} metric, let $S_i$ denote a nonnegative aggregate engagement measure for unit $i$ over the readout window, and define $A_i = \mathbb{I}(S_i > 0)$ as an indicator that unit $i$ exhibits any engagement. The group-level engagement depth estimand corresponds to the expected aggregate engagement conditional on positive engagement:
\begin{align}
    \mu_g^{\text{depth}} = \mathbb{E}\!\left[ S_i \mid i \in g, A_i = 1 \right],
\end{align}
with empirical estimator
\begin{align}
    \hat{\mu}_g^{\text{depth}} = \frac{\sum_{i=1}^{n_g} S_i}{\sum_{i=1}^{n_g} A_i}.
\end{align}

For both engagement breadth and engagement depth, the percentage lift is defined as the relative change of the test mean with respect to the control mean:
\begin{align}
    \widehat{\text{Lift}}(\%) = 100 \times \left( \frac{\hat{\mu}_{\text{test}}}{\hat{\mu}_{\text{control}}} - 1 \right).
\end{align}

Since the estimand is a ratio of means, symmetric normal approximations can be unreliable, particularly when the uncertainty of the denominator is non-negligible or when the distribution is skewed. We therefore construct confidence intervals (CIs)\,\citep{efron1994introduction} using Fieller's Theorem\,\citep{fieller1954some}. This method provides an exact (under normality) or near-exact CI for a ratio of estimators, yielding naturally asymmetric bounds in the original scale. We refer the reader to Appendix~\ref{appendix:ci} for details of these calculations.

\subsection{Safety and Privacy}
\label{sec:privacy_safety}

The CharacterFlywheel model development process operates under a rigorous framework of continuous monitoring and compliance with safety standards. We inherit safety standards from Llama 3.1~\citep{grattafiori2024llama} with the primary objective of minimizing safety violations and reducing false refusals, thereby promoting a system that prioritizes harmless outputs. We apply a layered system of automatic and manual evaluations, guided by critical safety rules, risk standards.

Key components of the system include:
\begin{itemize}
\item \textbf{Layered Evaluation}: Safety classifiers are performed automatically at multiple stages—during character auto-generation, character updates, model updates, user online interaction and user traffic sampling (Section~\ref{sec:curation}) of CharacterFlywheel model iterations. Manual human review is also triggered for difference scenarios, including uncertain characters, reported characters.
\item \textbf{Fail-Closed Design}: In the character creation phase, if any required safety rule fails, the system automatically rejects the creation, preventing unsafe characters from being published. Ambiguous cases are escalated to manual review for a final decision. During the user interaction phase, the system enforces safety by rejecting unsafe prompts and model responses automatically, ensuring all interactions comply with established safety standards. For model updates, only those models that successfully pass both automated evaluation and redteaming review are deployed to production.
\end{itemize}

Similar to safety measures, we apply privacy rules, including model-based and rule-based checks on high-risk identifiers, in the upstream data curation pipeline (Section~\ref{sec:curation}) to ensure no entity-identifiable information is included in downstream processing, including annotation and training.

\subsection{Image Generation}
\label{sec:img_gen_method}
Media generation is indispensable for modern social chat products, we incorporated image generation capabilities to support our primary objectives of providing social value to the user. Beyond standard utility, the novelty of this feature lies in leveraging visual content to actively enhance conversational engagement. Specifically, we developed image generation across two distinct scenarios:

\begin{itemize}
\item \textbf{Explicit Generation:} The user explicitly prompts the model to create an image, functioning similarly to standard multimodal chatbots.

\item \textbf{Implicit Generation:} A novel mechanism within CharacterFlywheel where the LLM autonomously decides to trigger image generation when it determines that visual content will enrich the conversation.
\end{itemize}

We formulate this as an agentic tool-calling task~\citep{schick2023toolformerlanguagemodelsteach}, where the LLM learns both when to trigger generation during the interaction and what generation parameters to provide, including the image prompt required by a standalone downstream text-to-image (T2I) model~\citep{dai2023emuenhancingimagegeneration}. Given the inherent subjectivity of the task, particularly for implicit image generation, data annotation presents a significant challenge. To address this, we enforce multi-review annotation to retain only high-consensus data. We specifically mandate agreement on two key dimensions: 1) the appropriateness of triggering generation at the current conversational turn, and 2) the quality of the image prompt in capturing the full conversational context, including history. Preference data are collected for training preference models with image generation, which are then used to construct the SFT and DPO datasets in the corresponding training stages.

\newpage

\section{Results}

From January 2024 to April 2025, we continuously iterated on CharacterFlywheel model development for a total of 15 versions, spanning both pre-launch (V1-V7) and post-deployment (V8-V15) phases. Section~\ref{sec:overall_quality_and_engagement} demonstrates the effectiveness of our iterative development framework in accumulating improvements over iterations, measured through online A/B engagement metrics, human win-rates, and reward model performance. Section~\ref{sec:community_benchmarks_and_steerability} presents our models' performance on standard LLM benchmarks (MMLU, GSM8K, MATH, HumanEval) and character instruction-following capabilities compared to Llama 3.1 baselines. Section~\ref{sec:responses_char_and_safety} analyzes the evolution of response patterns, formatting, sentiment, failure modes, violation rates, and false refusal rates across model versions. Section~\ref{sec:preference_modeling_results} examines the performance and behavior of our reward models. Together, these results demonstrate sustained improvements in user engagement while maintaining model quality, safety, and steerability across millions of real-world interactions.

\subsection{Quality and Engagement}
\label{sec:overall_quality_and_engagement}

\subsubsection{Pre-Launch Development and Validation}

\begin{figure}[ht]
    \centering
    \includegraphics[width=1\linewidth]{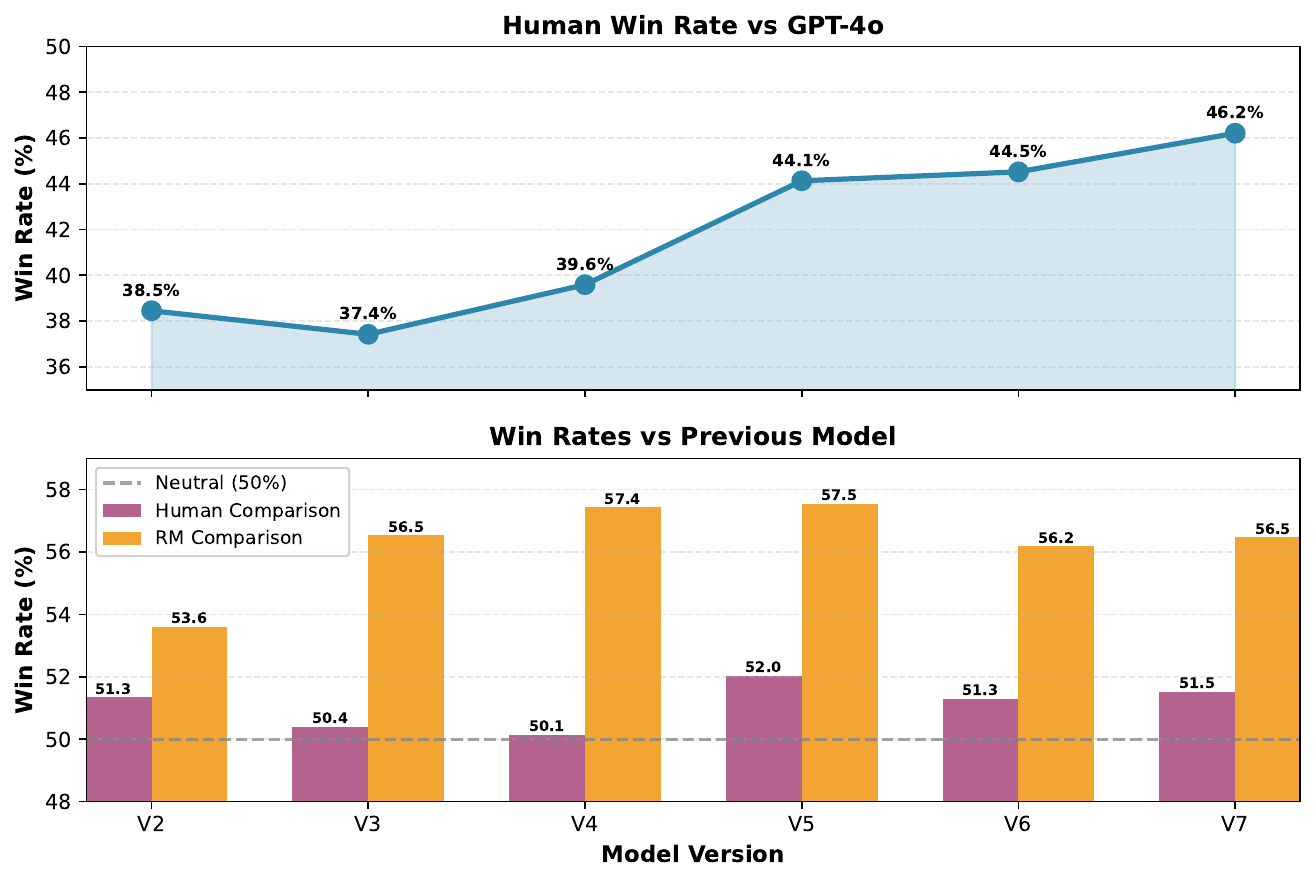}
    \caption{\textbf{Pre-launch quality progression (V2-V7).} \textit{Left panel:} Human win rates comparing CharacterFlywheel models against GPT-4o baseline, showing improvement from 37.4\% (V3) to 46.2\% (V7). \textit{Right panel:} Win rates against the immediate previous model version, displaying both human evaluations (blue bars, 50.2\%-52.5\%) and reward model predictions (orange bars, 53.6\%-57.6\%). Both signals consistently exceed the 50\% neutral threshold (dashed line), indicating quality improvements across iterations.}
    \label{fig:pre_lanuch_winrates}
\end{figure}

During the pre-launch phase from January 2024 to July 2024, we developed seven model versions (V1-V7) through iterative refinement, relying primarily on offline evaluation metrics to guide model selection. Figure~\ref{fig:pre_lanuch_winrates} demonstrates consistent quality improvements across successive versions. Human evaluations against GPT-4o show steady progression from 37.4\% (V3) to 46.2\% (V7), while both human and reward model win rates against previous versions consistently exceeded the 50\% neutral threshold, confirming progressive quality gains throughout pre-launch development.

\begin{wrapfigure}{r}{0.5\textwidth}
    \vspace{-10pt}
    \centering
    \includegraphics[width=0.48\textwidth]{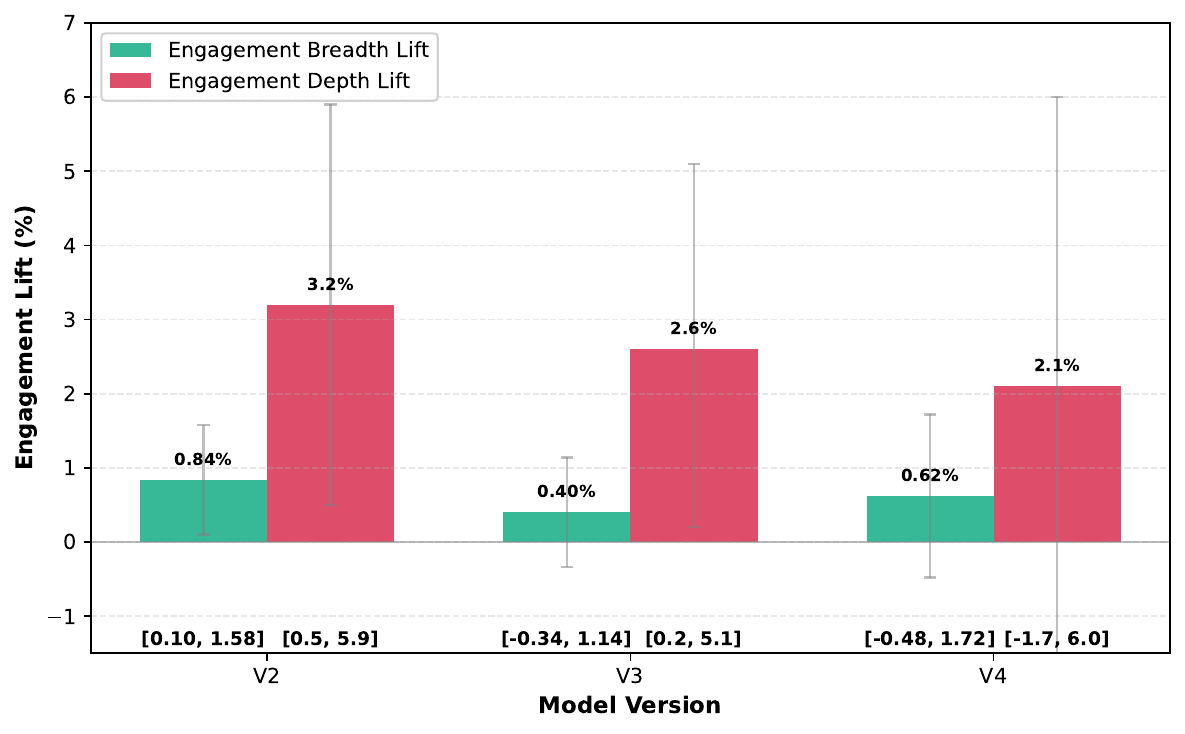}
    \caption{\textbf{Pre-launch engagement validation.} Engagement breadth and depth metric lifts with 95\% confidence intervals (Section~\ref{sec:online_eval}) for V2-V4, tested on a fixed set of characters with random online users.}
    \label{fig:pre_lanuch_engagement}
    \vspace{-10pt}
\end{wrapfigure}

To validate that offline quality improvements translated to online user engagement, we conducted small-scale A/B experiments on V2, V3, and V4. Figure~\ref{fig:pre_lanuch_engagement} shows consistent positive lift across both engagement breadth and depth metrics comparing to previous versions in A/B tests.
Despite limited statistical power in these early tests—with some confidence intervals including zero due to small sample sizes—the directional consistency provided early validation that our offline optimization approach aligned with online engagement objectives, giving us confidence to proceed toward full-scale launch in July 2024.

\subsubsection{Post-Launch Iterative Improvement}

\begin{figure}[ht]
    \centering
    \includegraphics[width=1\linewidth]{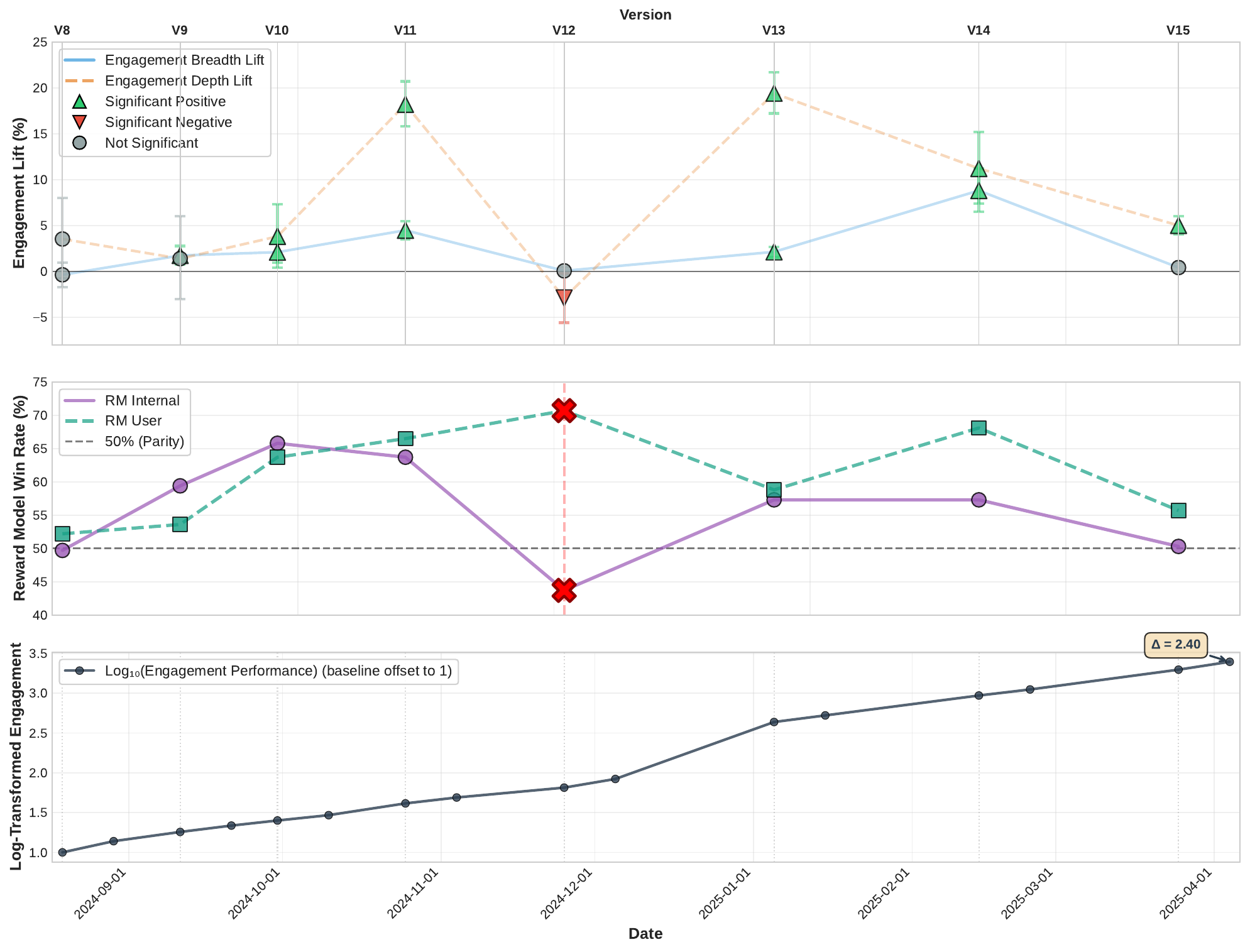}
    \caption{\textbf{Post-launch engagement trajectory (V8-V15).} \textit{Top panel:} A/B test engagement lifts comparing each version against the production baseline. Green markers indicate statistically significant positive results, red indicates significant negative, and gray indicates non-significant. \textit{Middle panel:} Reward model win rates over time for models trained on internal traffic (blue) and user traffic (orange). The spike at V12, where RM User exceeds 70\% while online engagement degrades, illustrates reward overfitting. \textit{Bottom panel:} Cumulative Engagement growth (baseline offset to 1) showing sustained upward trend of engagement though not entirely attributable to model updates.}
    \label{fig:engagement_trajectory}
\end{figure}

Following the July 2024 public launch, we continued iterative development through eight additional versions (V8-V15) deployed between August 2024 and April 2025. Figure~\ref{fig:engagement_trajectory} demonstrates sustained effectiveness across three complementary dimensions: A/B test lifts comparing new model versions against previous versions on engagement breadth and depth metrics, reward model win rates, and cumulative growth trajectory.

The majority of deployed versions achieved statistically significant positive engagement lifts, with notable successes including V11 (+4.47\% breadth, +18.2\% depth) and V14 (+8.8\% breadth, +11.2\% depth). However, we also encountered setbacks, most notably V12, which exhibited only +0.05\% breadth and -2.9\% depth. The V12 failure proved particularly instructive, revealing critical insights about reward model overfitting and safe optimization practices.

\textbf{Mitigating Overfitting.} The V12 degradation provides a cautionary example of reward model overfitting during RL optimization. As shown in the middle panel of Figure~\ref{fig:engagement_trajectory}, V12's RM win rate on user traffic spiked to 70.7\%—significantly higher than the typical 50-65\% range maintained by successful versions—while the RM win rate on internal traffic dropped to 43.7\%. This aggressive optimization pushed the policy into regions of the reward landscape where the reward model had low confidence, analogous to climbing beyond the reliable contours of our optimization map. The severe divergence between the two reward signals served as a critical warning that the model was overfitting to the user traffic distribution used for RM training rather than generalizing to broader real-world interactions.

This experience established important guardrails for our development process. We now monitor both RM Internal and RM User win rates as complementary signals, treating significant divergence or unusually high win rates as red flags for potential overfitting. Based on empirical observations across successful and failed deployments, we established a safer operating threshold: RM win rates should remain below 65\%, with 60\% being the ideal target for sustainable optimization. This threshold ensures we exploit the reward signal while maintaining sufficient margin from the unreliable regions of the learned reward landscape. Subsequent versions (V13-V15) adhered to these constraints and demonstrated restored positive engagement trends, validating the effectiveness of this more conservative approach.

Despite V12 setbacks, the cumulative engagement trajectory shows a clear upward trend, demonstrating substantial accumulated engagement gains though not entirely attributable to model updates. Together, these results validate our core hypothesis: by tightly integrating human feedback, reward modeling, and online A/B testing within a robust iteration loop—while maintaining appropriate safeguards against overfitting—we can achieve consistent, measurable improvements in user engagement for social conversational applications at scale.

\subsection{Community Benchmarks and Steerability}
\label{sec:community_benchmarks_and_steerability}
While CharacterFlywheel is optimized primarily for engagement in social chat scenarios rather than utility tasks, we evaluate performance on standard community benchmarks to ensure the model maintains robust general capabilities. Additionally, we assess the model's ability to adhere to character-specific instructions—a important capability for our application where users create customized AI personas.
\subsubsection{Performance on Standard Benchmarks}
Table~\ref{tab:ch_v7_benchmarks} presents CharacterFlywheel V7's performance across widely-used LLM benchmarks, comparing against Llama 3 family models and leading commercial systems. CharacterFlywheel V7, fine-tuned from Llama 3.1 70B, maintains competitive performance across most benchmarks despite being optimized for social engagement rather than pure utility.
\begin{table}[h]
\centering
\small
\begin{tabular}{@{}llc!{\vrule width 0.8pt}cc!{\vrule width 0.8pt}cccc@{}}
\toprule
\rotatebox{00}{\makebox{\textbf{Category}}} & \rotatebox{00}{\makebox{\textbf{Benchmark}}} & \rotatebox{90}{\makebox{\textbf{Llama 3 8B}}} & \rotatebox{90}{\makebox{\textbf{Llama 3 70B}}} & \rotatebox{90}{\makebox{\textbf{CharacterFlywheel V7}}} & \rotatebox{90}{\makebox{\textbf{Llama 3 405B}}} & \rotatebox{90}{\makebox{\textbf{GPT-4o}}} & \rotatebox{90}{\makebox{\textbf{Claude 3.5 Sonnet}}} \\
\midrule
\textbf{General} & MMLU (0-shot) & 60.4 & \cellcolor{lightgray!50}83.6 & \cellcolor{lightgray!50}79.5 & 87.3 & 89.1 & 89.9 \\
 & IFEval & 84.4 & \cellcolor{lightgray!50}87.5 & \cellcolor{lightgray!50}84.8 & 88.6 & 85.6 & 88.0 \\
\midrule
\textbf{Code} & HumanEval (0-shot) & 72.6 & \cellcolor{lightgray!50}80.5 & \cellcolor{lightgray!50}77.4 & 89.0 & 90.2 & 92.0 \\
 & MBPP (3-shot CoT) & 71.7 & \cellcolor{lightgray!50}86.0 & \cellcolor{lightgray!50}66.6 & 88.6 & 87.8 & 90.5 \\
\midrule
\textbf{Math} & GSM8K (8-shot CoT) & 84.5 & \cellcolor{lightgray!50}95.1 & \cellcolor{lightgray!50}92.3 & 96.8 & 96.1 & 96.4 \\
 & MATH (0-shot CoT) & 51.9 & \cellcolor{lightgray!50}68.0 & \cellcolor{lightgray!50}50.5 & 73.8 & 76.6 & 71.1 \\
\midrule
\textbf{Reasoning} & ARC Challenge (0-shot) & 83.4 & \cellcolor{lightgray!50}94.8 & \cellcolor{lightgray!50}93.1 & 96.9 & 96.7 & 96.7 \\
 & GPQA (0-shot CoT) & 32.8 & \cellcolor{lightgray!50}46.7 & \cellcolor{lightgray!50}39.3 & 51.1 & 53.6 & 59.4 \\
\bottomrule
\end{tabular}
\caption{\textbf{Benchmark comparison.} CharacterFlywheel V7 performance compared to Llama 3 family and leading commercial models. CharacterFlywheel V7 is fine-tuned from Llama 3 70B (highlighted baseline).}
\label{tab:ch_v7_benchmarks}
\end{table}
On general knowledge and instruction-following tasks, CharacterFlywheel V7 achieves 79.5\% on MMLU and 84.8\% on IFEval, representing modest degradation from the Llama 3 70B baseline (83.6\% and 87.5\% respectively) but remaining competitive with other leading models. Mathematical reasoning shows strong retention with 92.3\% on GSM8K (vs. 95.1\% baseline), though performance on the more challenging MATH benchmark drops to 50.5\% from 68.0\%. Coding performance exhibits the most significant gap, with HumanEval at 77.4\% (vs. 80.5\%) and MBPP at 66.6\% (vs. 86.0\%). Reasoning benchmarks remain robust, with ARC Challenge at 93.1\% (vs. 94.8\%) and GPQA at 39.3\% (vs. 46.7\%).
\begin{figure}[ht]
    \centering
    \includegraphics[width=0.95\linewidth]{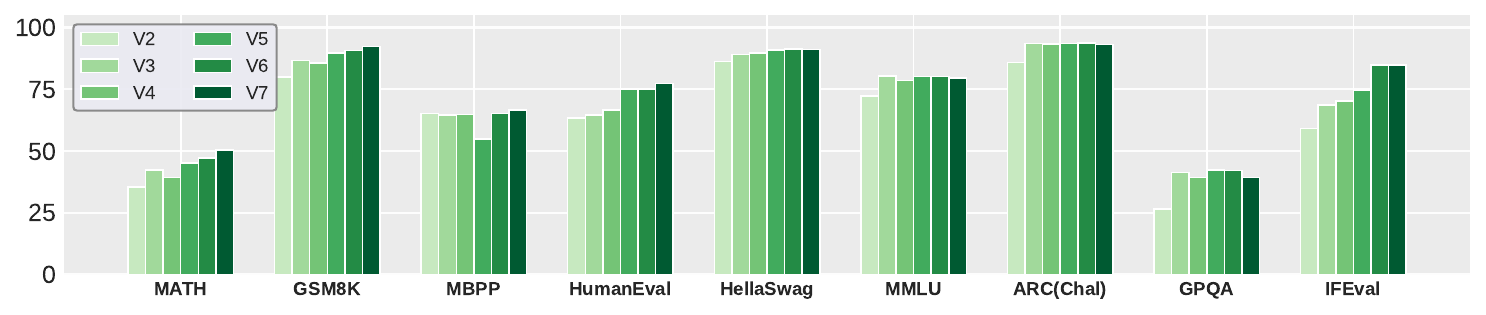}
    \caption{\textbf{Benchmark progression across pre-launch versions (V2-V7).} Performance across nine standard benchmarks shows general stability throughout development, with most metrics converging by V6-V7. Notable improvements occur in IFEval and ARC Challenge, while MBPP shows some degradation, reflecting the trade-offs in optimizing for conversational engagement.}
    \label{fig:benchmark_progression}
\end{figure}
Figure~\ref{fig:benchmark_progression} illustrates benchmark evolution throughout pre-launch development. Most benchmarks remained stable or improved across versions, with particularly notable gains in IFEval (climbing from approximately 75\% in V2 to 84.8\% in V7) and ARC Challenge (rising to 93.1\%). The relative stability across iterations validates that our engagement-focused optimization did not catastrophically degrade general capabilities—the model retains sufficient versatility to handle diverse query types encountered in production, from factual questions to casual conversation.
\subsubsection{Character Steerability}

A critical capability for CharacterFlywheel is \textit{character steerability}—the model's ability to consistently adhere to user-specified character instructions and personas throughout conversations. Unlike IFEval, which tests verifiable task completion, character steerability requires maintaining personality traits, tone, and behavioral patterns in open-ended social interactions.

\begin{table}[htbp]
\centering
\small
\begin{tabular}{@{}lccccccc@{}}
\toprule
\textbf{Version} & V2 & V3 & V4 & V5 & V6 & V7 & V8 \\
\midrule
\textbf{Instruction Violation (\%)} & 26.6 & 22.6 & 17.9 & 22.2 & 13.7 & 7.5 & 5.8 \\
\bottomrule
\end{tabular}
\caption{\textbf{Character steerability improvement.} Instruction violation rates measured via LLM-as-a-judge on interactive chat.}
\label{tab:instruction_violation}
\end{table}

We measure steerability through two metrics: (1) IFEval performance for general instruction-following, and (2) instruction violation rate for character-specific adherence. CharacterFlywheel V7 achieves 84.8\% on IFEval, maintaining competitive performance with the Llama 3 70B baseline (87.5\%) and other leading models.

Table~\ref{tab:instruction_violation} shows dramatic improvement in character adherence, with instruction violations decreasing from 26.6\% (V2) to 5.8\% (V8)—a 78\% relative reduction. We evaluate this metric using LLM-as-a-judge on interactive chat sessions where internal users and domain experts conduct mildly adversarial conversations, naturally testing the model's ability to maintain character instructions under realistic but challenging conditions. Violations include refusing to discuss appropriate topics, adopting incorrect tone, or failing to maintain specified personality traits.

This improvement emerged naturally from our annotation process (Section~\ref{sec:data_annotation}). While annotators primarily focused on engagement quality, they also labeled instruction violations and edited responses when necessary to better align with character instructions. This dual-focus approach demonstrates that optimizing for engagement and instruction-following need not be in conflict.

\subsection{Response Characteristics}
\label{sec:responses_char_and_safety}

\begin{figure}[ht]
    \centering
    \includegraphics[width=1\linewidth]{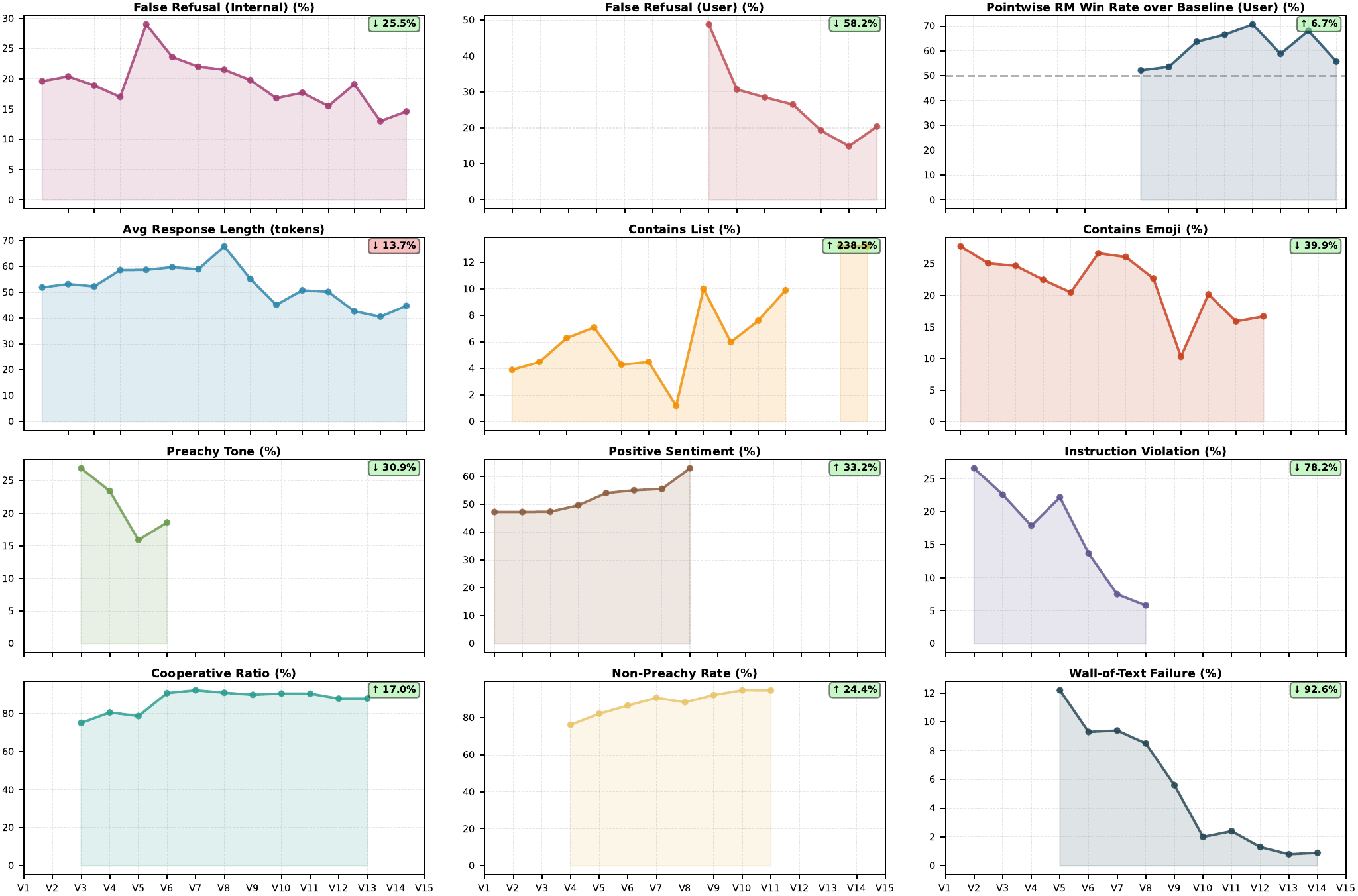}
    \caption{\textbf{Response characteristics across model versions (V1-V15).} Evolution of false refusal metrics (top row), formatting patterns (second row), tone and sentiment (third row), and quality indicators (bottom row). Percentage labels indicate relative change from V1 to V15. Metrics are evaluated using LLM-as-a-judge on internal annotated traffic and rule-based methods on user traffic evaluation sets.}
    \label{fig:response_characteristics}
\end{figure}

Beyond engagement and benchmark performance, we track a comprehensive suite of response characteristics and safety metrics throughout development to ensure model quality, identify failure modes, and maintain safety standards. Figure~\ref{fig:response_characteristics} presents the evolution of twelve key metrics across all 15 model versions, measured using a combination of LLM-as-a-judge and rule-based methods on curated evaluation sets (Section~\ref{sec:online_eval}).

A critical observation from Figure~\ref{fig:response_characteristics} is the \textit{robustness of the iterative optimization process}. While individual versions occasionally exhibit spikes in undesirable metrics—such as elevated false refusals in V5-V6—these degradations are consistently smoothed out within a few subsequent iterations when efficient solutions and robust monitoring are in place. This self-correcting behavior demonstrates that our iterative framework, when coupled with comprehensive metric tracking, naturally navigates toward improved quality even when individual optimization steps temporarily overshoot or introduce new failure modes. The overall downward trends in negative metrics and upward trends in positive metrics validate that the development process is fundamentally stable and convergent.

\subsubsection{False Refusal}

\textbf{False refusal} metrics track instances where the model unnecessarily declines benign requests. On internal traffic, false refusals show considerable variation across versions, with peaks at V5-V6 exceeding 30\%, but demonstrate overall 25.5\% improvement, settling around 20\% by V15. On user traffic, we observe even more dramatic improvement, with false refusals dropping from over 20\% to under 5\%, despite temporary increases in mid-development versions. These fluctuations and recoveries demonstrate that the iterative process can absorb and correct over-cautious safety tuning—when a version becomes too conservative, subsequent iterations recalibrate based on false refusal signals, restoring appropriate balance.

\subsubsection{Response Formatting and Style}

We monitor several formatting and stylistic characteristics to ensure responses remain natural and appropriately varied. \textbf{Average response length} remained relatively stable throughout development, ranging from approximately 50-65 tokens, with controlled variations reflecting different optimization priorities across versions. \textbf{Contains List} percentage shows moderate fluctuation (13.7\% relative change), indicating the model maintains flexibility in using structured formatting when appropriate without over-relying on lists.

\textbf{Contains Emoji} exhibits the most dramatic variation (238.5\% relative change), with notable spikes at V11-V12. While emojis can enhance conversational tone and engagement, excessive use appears templated and inauthentic. The sharp increase followed by subsequent normalization exemplifies the self-correcting nature of our process: when V12's aggressive emoji usage became apparent through both this metric and engagement degradation, we adjusted annotation guidelines and data composition in V13-V15, successfully moderating emoji frequency to more appropriate levels.

\subsubsection{Tone and Sentiment Quality}

Tone-related metrics reveal significant improvements in conversational quality. \textbf{Preachy tone}, which measures overly moralizing or lecturing responses, decreased 30.9\% from approximately 25\% to under 18\%, despite some mid-iteration peaks. This reduction addresses a common complaint in AI assistants that adopt condescending or judgmental tones. Conversely, \textbf{positive sentiment} increased 33.2\%, rising from approximately 45\% to 60\%, indicating that responses became more uplifting and supportive without becoming preachy.

The \textbf{cooperative ratio}, measuring the model's willingness to assist versus unnecessary resistance, improved dramatically (78.2\% relative change), climbing from under 60\% to over 80\%. Similarly, \textbf{non-preachy rate} reached 92.6\%, indicating the vast majority of responses avoid moralizing tone. Notably, both metrics show temporary dips at certain versions (particularly V5-V6 for cooperative ratio), which were subsequently recovered, reinforcing that monitored iteration naturally corrects for over-tuning in either direction.

\subsubsection{Quality and Failure Modes}

\textbf{Instruction violation}, already discussed in Section~\ref{sec:community_benchmarks_and_steerability}, shows a 39.9\% reduction from approximately 27\% to 16\% when measured on this broader evaluation set. \textbf{Wall-of-text failure}, measuring responses with poor formatting and excessive length without structure, decreased 58.2\% from over 10\% to under 5\%. This improvement enhances readability and user experience, particularly on mobile devices where large unformatted text blocks are difficult to parse.

Together, these metrics demonstrate that our iterative development successfully improved not only engagement but also response quality, safety, and conversational tone. Crucially, the dynamics reveal that temporary metric degradations—inevitable in any optimization process—are reliably resolved within a few iterations when comprehensive monitoring is maintained. This robustness emerges from our feedback loop design: when a version exhibits elevated failure rates or degraded characteristics, these signals immediately inform the next round of data annotation and model training, enabling rapid course correction. The general convergence across metrics from V1 to V15, despite intermediate volatility, validates that properly monitored iterative optimization is a stable and effective approach for improving complex, multi-objective systems like social conversation.

\begin{table}[h]
\centering
\footnotesize
\caption{Failure modes and mitigation strategies.}
\label{tab:user_signals}
\begin{tabular}{p{3cm}p{4.5cm}p{6.5cm}}
\toprule
\textbf{Failures} & \textbf{Description} & \textbf{Mitigation strategies} \\
\midrule
Infinity Loop & The model occasionally generates repetitive responses, resulting in excessively long outputs that exceed the token limit. & Employed token count as a heuristic to identify prompts that trigger repetitive responses; Masked EOT tokens in DPO and online DPO loss computation; Introduced a long response penalty in online RL. Led to a 57.8\% lower loop rate in online traffic. \\
\midrule
Repetitive Phrases & The model over-uses common phrases, such as "OMG" (4.14\%), "LOL" (3.3\%), "mind blown" (3.2\%), and templated refusal. Moreover, markdown formatting in responses is inconsistent (e.g., 37\% unclosed or broken formatting). & Verified that the model has a tendency to follow undesirable “biases” in prompts (Sec. \ref{sec:biases-piror-turns}). Developed a linter to remove or reduce the prevalence of repetitive phrases and to enforce consistent Markdown formatting in SFT data and online RL prompts. Reduced repetitive phrases by 60\% in online traffic. \\
\midrule
Hindi Language Mismatches & The model responds in English to users who chat in Hindi and sometimes provides unrequested translations to English. & Curated SFT and DPO datasets based on failures in online traffic; Collected extra preference pairs to correct the RM’s bias towards English responses in Hindi prompts. Reduced LMR failures from 75\% to 1.5\%, and the prevalence of implicit translation from 64.4\% to 1.1\% on a hard evaluation set. \\
\bottomrule
\end{tabular}
\end{table}

\subsection{Preference Modeling}
\label{sec:preference_modeling_results}
\begin{table}[htbp]
\centering
\caption{Pointwise Preference Model Performance Across Different Data Batches.}
\label{tab:rm_results}
\resizebox{\textwidth}{!}{%
\begin{tabular}{@{}l*{5}{c}*{5}{c}@{}}
\toprule
& \multicolumn{5}{c}{\textbf{Static Chat Preference Data}} & \multicolumn{5}{c}{\textbf{Interactive Chat Preference Data}} \\
\cmidrule(lr){2-6} \cmidrule(lr){7-11}
\textbf{Model} & Batch & Batch & Batch & Batch & All Data & Batch & Batch & Batch & Batch & All Data \\
& 240923 & 241018 & 241118 & 241229 & until 241229 & 240923 & 241018 & 241118 & 241229 & until 241229 \\
\midrule
RM\_240923 & \textbf{0.725} & 0.572 & 0.558 & 0.575 & 0.652 & \textbf{0.608} & 0.503 & 0.500 & 0.513 & 0.573 \\
RM\_241018 & 0.715 & \textbf{0.721} & 0.557 & 0.555 & 0.672 & 0.634 & \textbf{0.644} & 0.510 & 0.521 & 0.627 \\
RM\_241118 & 0.732 & 0.743 & \textbf{0.725} & 0.619 & 0.720 & 0.645 & 0.624 & \textbf{0.628} & 0.546 & 0.632 \\
RM\_241229 & 0.742 & 0.735 & 0.754 & \textbf{0.742} & \textbf{0.746} & 0.644 & 0.631 & 0.648 & \textbf{0.669} & \textbf{0.650} \\
\bottomrule
\end{tabular}%
}
\end{table}
We adopt an iterative approach to preference modeling, where we incrementally incorporate new training data through our data pipeline at each iteration. Table~\ref{tab:rm_results} presents the performance of reward models trained at different stages of this iterative process. The version number of each reward model (RM\_YYMMDD) indicates the latest batch of data included in its training set. For example, RM\_241018 was trained on all data available up to and including the October 18, 2024 batch.

\textbf{Zero-shot generalization to unseen batches.} 
We first examine how reward models generalize to future, unseen data batches. The off-diagonal entries in Table~\ref{tab:rm_results} reveal that models exhibit moderate zero-shot performance on subsequent batches, achieving approximately 55-60\% accuracy on static chat preference data and 50-55\% on interactive chat preference data. This near-random performance on interactive data suggests a higher degree of distribution shift or task complexity compared to static preferences, which we attribute to the dynamic nature of interactive conversations.

\textbf{Performance gains on newly incorporated batches.}
When a reward model is trained with a new data batch, we observe consistent performance improvements on that specific batch (highlighted diagonal entries). For instance, RM\_241229 achieves 0.742 accuracy on Batch\_241229 for static data and 0.669 for interactive data, representing substantial gains over the prior model's zero-shot performance (RM\_241118: 0.619 and 0.546, respectively). Critically, these improvements come with minimal regression on previously seen batches. Comparing the performance of successive models on earlier batches (e.g., RM\_240923 vs. RM\_241018 vs. RM\_241118 on Batch\_240923), we observe stable or slightly improved scores, indicating that the model successfully retains knowledge from earlier data while adapting to new distributions.

\textbf{Limited backward transfer.}
Interestingly, we do not observe significant improvements on previous batches when training with additional data. For example, RM\_241229's performance on Batch\_240923 (0.742 static, 0.644 interactive) remains comparable to RM\_240923's original performance (0.725 static, 0.608 interactive). This suggests that each batch captures relatively distinct preference patterns, and newer data does not substantially refine the model's understanding of earlier distributions.

\textbf{Evidence of consistent preference quality accumulation.}
The overall performance on the aggregated "All Data until 241229" evaluation set shows steady improvement across iterations: from 0.652 (RM\_240923) to 0.746 (RM\_241229) for static data, and from 0.573 to 0.650 for interactive data. This progressive improvement, coupled with the absence of catastrophic forgetting or inter-batch conflicts, validates the success of our iterative framework for accumulating preference quality. We attribute this compatibility between batches to our deliberate decision to maintain consistent annotation guidelines throughout the data collection process, adjusting minimally only when necessary to accommodate evolving product dynamics. This consistency ensures that preference labels across different time periods remain aligned, enabling effective knowledge accumulation without introducing conflicting supervision signals.

\subsubsection{Impact of Annotation Agreement on Engagement Preference Modeling}
To investigate the impact of annotation agreement on engagement preference modeling, we conducted a controlled experiment using a subset of preference pairs sampled from user traffic. Each pair was assessed by three independent annotators. From these annotations, we constructed three training data variants:
\begin{itemize}
    \item \textbf{Multi-Review (With Agreement):} Includes only data points with unanimous agreement across all three annotators.
    \item \textbf{Single-Review (All):} Includes all annotations, retaining conflicting labels for the same data point.
    \item \textbf{Single-Review (Random):} Includes each data point once, with a label randomly sampled from the three annotators.
\end{itemize}

We split the dataset and trained Pointwise and Pairwise reward models (Section~\ref{sec:pref-rm}) on each variant and evaluated their performance on both \textbf{Single-Review (Random)} and \textbf{Multi-Review (With Agreement)} evaluation sets. The results are shown in Table~\ref{tab:multi_review_results}.

\begin{table}[ht]
\centering
\small
\caption{Impact of annotation agreement on engagement preference accuracy. We compare a baseline (Not Trained) against models trained on different data variants, using two distinct evaluation sets.}
\label{tab:multi_review_results}
\begin{tabular}{lcccc}
\toprule
\multirow{3}{*}{\textbf{Training Data}} & \multicolumn{2}{c}{\textbf{Pointwise RM Acc (\%)}} & \multicolumn{2}{c}{\textbf{Pairwise RM Acc (\%)}} \\
\cmidrule(lr){2-3} \cmidrule(lr){4-5}
 & \multicolumn{2}{c}{\textbf{Evaluation Set}} & \multicolumn{2}{c}{\textbf{Evaluation Set}} \\
 & \textit{Single (Rand)} & \textit{Multi (Agree)} & \textit{Single (Rand)} & \textit{Multi (Agree)} \\
\midrule
\textit{Baseline (Not Trained)} & 59.01 & 66.84 & 59.10 & 67.08 \\
\midrule
Multi-Review (With Agreement) & 59.02 & \textbf{70.86} & 60.60 & \textbf{72.87} \\
Single-Review (All) & 58.94 & 70.59 & \textbf{61.50} & 72.08 \\
Single-Review (Random) & 60.18 & 70.96 & 60.70 & 72.50 \\
\bottomrule
\end{tabular}
\end{table}

Our experiments reveal two important insights. \textbf{Multi-review evaluation is essential for measuring model improvement.} When comparing trained models to untrained baselines, performance on the Single-Review (Random) set remains nearly unchanged. However, we observe substantial improvement when evaluating on the Multi-Review (With Agreement) set (e.g., +4.02 points for Pointwise RM). This discrepancy indicates that single-review benchmarks contain too much label noise to reliably distinguish trained models from baselines, whereas consensus-based evaluation provides a stable ground truth for assessment. \textbf{Single-review data can effectively guide model training.} We find that models trained on single-review data—whether using all conflicting labels or random sampling—still achieve strong performance on the rigorous Multi-Review evaluation. This suggests that the model can often distill robust preference patterns by aggregating noisy signals from diverse perspectives. \textit{However}, we note that our experiments focused on engagement quality, which is inherently subjective. While lower-agreement data proved effective in this context, tasks requiring high objective utility or factual precision may still necessitate stricter agreement standards.

\subsection{Analysis \& Discussion}

\subsubsection{The impact of Image Generation}
We started public feature of explicit image generation in V9, achieving +1.7\% Engagement Breadth Metric Lift over text-only baselines in a 7-day A/B tests. In V10, implicit image generation became the primary engagement driver, contributing to an additional +2.1\% Engagement Breadth Metric Lift over V9 (alongside other improvements). This highlights the value of autonomous image generation that enriches conversations without requiring explicit user prompts.

\subsubsection{On-policy vs. Off-policy}
\label{sec:on_policy_off_policy}

We compared two RL prompt sets by training the same starting checkpoint: (a) prompts from the latest model's traffic (near-policy), and (b) prompts from earlier model versions. The near-policy set achieved significant advantages: +10.6\% Engagement Depth Metric Lift and +1.6\% Engagement Breadth Metric Lift compared to the off-policy counterpart in an A/B test.
This result aligns with our engagement landscape navigation framework in Section~\ref{sec:methodology_development_cycle}: we obtain the most effective gradients when training samples sufficiently estimate the contour near the current policy, enabling continual hill-climbing within the policy space.

\subsubsection{Online DPO vs. GRPO}
\label{sec:analysis_discussion_odpo_vs_grpo}
We conducted an A/B test comparing two models initialized with the same checkpoint and training data, but trained with either Online DPO loss or GRPO loss. The results show that the model trained with GRPO loss achieved a +1.52\% Engagement Breadth Metric Lift over the model trained with Online DPO loss. This is likely due to its ability to exploit reward scores from all generated responses, which provide a more fine-grained supervision signal.

\subsubsection{Variance-based Downsampling}
\label{sec:variance-based-downsampling}

A key principle in post-training is focusing on hard prompts where the latest model struggles~\citep{yu2025rip}. A standard heuristic selects prompts with the lowest average RM scores. However, we find this unreliable in our study: preference RMs do not regularize score magnitudes across prompts, so scores often reflect stylistic factors (e.g., length, conversation turns) rather than difficulty. For instance, longer-turn conversations receive systematically lower scores, causing 4$\times$ over-representation of Roleplay and Romantic prompts when sampling based on average RM scores.

We instead use a \textbf{variance-based} strategy: computing RM score variance across multiple responses per prompt and prioritizing high-variance prompts. Difficult prompts induce wider spreads in response quality, making variance a more robust difficulty signal than mean score.

\subsubsection{User Signal Models}
\label{sec:analysis_discussion_user_signal_models}
In a small scale study, we found that $p(\text{continue})$ and $p(\text{thumb up})$ have high correlation with our preference reward model, showing similar win-rates between models and an upward scoring trend from CH7-12. However, we learned that user signal models are unsuitable for direct optimization in online RL due to several inherent biases that make them susceptible to reward hacking:

\textbf{Delayed feedback:} We observed that users often skip thumbs-up on early clarifying responses but react to final concrete answers. This causes P-Thumbs models to favor verbose responses over clarification questions.

\textbf{Ending bias:} We found users typically thumbs-up at conversation end where flattery (``thank you,'' ``have a good night'') is common, reproducing the sycophancy issue identified in ChatGPT~\cite{openai2024sycophancy}.

\textbf{Inconsistent pos/neg ratios:} We discovered that positive-to-negative label ratios vary significantly across JTBDs (e.g., ``Role-playing -- Romantic'' vs. ``Image-gen''), enabling RMs to shortcut by detecting JTBDs rather than assessing quality.

\textbf{Confounding context:} We learned that RMs over-index on prior-turn sentiment or satisfaction signals rather than focusing on last-turn response quality.

Given these limitations and failure of the V12 experiment, we use user signal models as additional scores in rejection sampling ranking rather than for direct RL optimization. This strategy effectively prevents over-optimization by constraining RM win-rates to below 70\%.

\subsubsection{Biases from Prior Turns}
\label{sec:biases-piror-turns}

Undesirable biases in online RL (e.g., excessive lists or emojis) are often attributed to the RM, but conversation history is an overlooked source. Since autoregressive policy models strongly mimic prior turn styles, fixing undesirable behaviors requires addressing them in training prompts, not just the RM.

We verified this with an emoji reduction experiment. Despite removing all emojis from RM inputs during scoring (i.e., "debiasing" the RM), average emoji count still increased from 0.2 to 0.48 over 120 RL steps. This confirms that models can inherit and amplify biases directly from conversation history. Consequently, we implement prompt pre-processing and enforce bias monitoring and mitigation as detailed in Section~\ref{sec:artifacts_mitigation}.





\subsection{Related Work}

\textbf{Character Chatbots and Conversational AI.} Research on conversational AI has deep historical roots, evolving from modular architectures like Xiaoice~\citep{zhou2020design} to modern transformer models with early adopters like Meena~\citep{adiwardana2020towards} and Blenderbot~\citep{roller2021recipes, shuster2022blenderbot}.
With the widespread adoption of LLMs in production systems, recent studies have focused on evaluating various dimensions of their social capabilities, including assessments of social intelligence~\citep{socialeval, gandhi2023understanding}, empathy~\citep{paech2023eqbench,phang2025investigating}, and companionship behavior~\citep{zhang2025dark,de2025ai}.
While commercial character chatbots such as Character.AI and Replika have reportedly gained substantial user adoption, their design and methodologies remain opaque. CharacterFlywheel pushes towards scientific rigor in conversationalist LLM development by detailing pipelines, training procedures, evaluation methods, and the rationale behind production-level deployment choices.

\textbf{Learning from Human Feedback and Organic Interaction.} The predominant paradigm in aligning LLMs involves Reinforcement Learning from Human Feedback (RLHF), typically utilizing carefully designed and crowdsourced datasets of pairwise preferences~\citep{ouyang2022training,bai2022training}. A complementary line of work focuses on harvesting training signals from organic interactions, such as conversation logs~\citep{hancock2019learning}, binary feedback like thumbs-up/down~\citep{xu2023learning,xu2023improving}, user message classifiers~\citep{chen2025retrospective,don2024naturally}, response length heuristics~\citep{pang2024leveraging}, comments~\citep{jaques2020human}, or immediate turn-level feedback~\citep{jin2025era}.
Our work emphasizes the nuanced integration of user conversations, implicit signals, and expert annotations within a robust framework that iteratively improves production engagement metrics

\textbf{Iterative LLM Refinement.} Early work demonstrated the possibility of lifelong continuous model improvement through deployment-based learning~\citep{shuster2020deploying}. Subsequently, a series of works on iterative preference optimization approaches—such as self-improvement~\citep{yuan2024self,rosset2024direct}, self-play~\citep{wuself}, and Nash policy optimization~\citep{zhangiterative}—have demonstrated superior convergence through repeated policy updates~\citep{xiong2024iterative}. Our work presents an online-product-metrics-centric perspective, exploring how differentiable reward models, offline reward model evaluation, and other training signals can collaboratively drive iterative improvement of non-differentiable online engagement metrics for conversational systems deployed in commercial applications serving millions of users.

\section{Conclusion}

This report presents CharacterFlywheel, a production-scale large language model optimized for social engagement across Instagram, WhatsApp, and Messenger. Through 15 iterative development cycles spanning 15 months, we demonstrate that systematic optimization for engagement, steerability and safety is both feasible and effective when combined with comprehensive monitoring and careful safeguards against overfitting. Our framework achieved sustained engagement improvements with cumulative engagement growth exhibiting a strong upward trend over nine months of post-launch deployment, while simultaneously improving steerability (78\% reduction in instruction violations).

We introduce several methodological contributions for developing engagement-focused conversational AI. Our iterative preference modeling approach demonstrates consistent engagement gain accumulation, enabled by maintaining consistent annotation guidelines. Critically, the V12 failure case revealed the dangers of reward model overfitting—when RM win rates spiked to 70.7\% while engagement degraded, we established empirical safeguards that subsequent versions should maintain RM win rates below 65\%. We also demonstrate that the iterative optimization process exhibits self-correcting behavior: temporary degradations in individual metrics (such as elevated false refusals, preachy tone, or emoji overuse) are reliably resolved within subsequent iterations through our comprehensive feedback loop.

Our work addresses a significant gap in LLM research by demonstrating that measurable progress is achievable even when optimizing for inherently subjective objectives like "engagingness." The sustained improvements across 15 iterations at production scale—serving millions of users with diverse use cases and languages—validate that systematic engineering approaches can succeed in domains traditionally considered too subjective for rigorous optimization. However, several challenges remain, including developing principled methods for detecting reward hacking, improving multi-turn optimization formulations, and better understanding of generalization in preference modeling.

We release this technical report to contribute to scientific progress in social LLM development and to provide practical guidance for teams building engagement-focused AI. The techniques presented here—particularly around reward model monitoring, preference data quality, and multi-metric evaluation—offer a template for developing and monitoring complex AI systems where single-metric optimization would be insufficient. As conversational AI continues expanding into social and entertainment applications, rigorous approaches to measuring and improving engagement while maintaining safety and quality will become increasingly critical.

\newpage

\section*{Acknowledgement}
We would like to express our sincere gratitude to the following individuals:
(a) Engineering Team: Grigorios Antonellis, Li Chen, Nikhil Dhanda, Asish Ghoshal, Zhaojie Gong, Fernando Gonzalez Adauto, Srikanth Grandhe, Matt Hall, Derek Hao Hu, Qiaoying Huang, Saeed Jahed, Yuchen Jiang, Slava Kachanovsky, Kimberly Kao, Vaibhav Kumar, Vinay Satish Kumar, Dongheng Li, Haoran Li, Yu-Hsiang Lin, Shubham Modi, Wancen Mu, Shrikant Nagori, Aasish Pappu, Chirag Parmar, Stan Peshterliev, Romil Shah, Tianyu Song, Krishna Pramod Sripada, Ye Tian, Desi Wang, Shaofei Wang, Pengxiang Wu, Yunfan Ye, Boer Zhang, Gelin Zhou, and Kangyuan Zhu for their contributions. (b) Engineering Management: Yang Bai, Ankit Jain, David Liu, Bowen Meng, Ketong Wang, Bo Xiong, and Anji Yi for their leadership throughout the project lifecycle. (c) Product Management: Neena Kamath and Sudip Shah for their product vision and guidance. (d) Leadership: Christine Awad, Jason Brewer, Wen Shi, and Kevin Tang for their strategic support and leadership. (e) Special Thanks: Jiabo Hu for developing the annotation UI for this project and Shruthi Patchava for sourcing annotation vendors.

\clearpage
\newpage
\bibliographystyle{assets/plainnat}
\bibliography{paper}

\clearpage
\newpage
\beginappendix

\section{Evaluation Performance Across Versions}
Table~\ref{tab:version_metrics} shows all evaluation results across 15 different CharacterFlywheel model versions.

\begin{table*}[h]
\centering
\small
\caption{Performance metrics across model versions}
\label{tab:version_metrics}
\resizebox{\textwidth}{!}{
\begin{tabular}{lccccccccccccccc}
\toprule
\textbf{Metric} & \textbf{V1} & \textbf{V2} & \textbf{V3} & \textbf{V4} & \textbf{V5} & \textbf{V6} & \textbf{V7} & \textbf{V8} & \textbf{V9} & \textbf{V10} & \textbf{V11} & \textbf{V12} & \textbf{V13} & \textbf{V14} & \textbf{V15} \\
\textbf{Release Date} & \textbf{Mar 11} & \textbf{Mar 28} & \textbf{Apr 23} & \textbf{May 28} & \textbf{Jun 17} & \textbf{Jul 10} & \textbf{Jul 22} & \textbf{Aug 19} & \textbf{Sep 11} & \textbf{Sep 30} & \textbf{Oct 25} & \textbf{Nov 25} & \textbf{Jan 5} & \textbf{Feb 14} & \textbf{Mar 25} \\
\midrule
\multicolumn{16}{l}{\textit{\textbf{User Engagement Metrics}}} \\
Engagement Breadth Lift (1P) (\%) & -- & \cellcolor{green!50}$0.84^{+0.74}_{-0.74}$ & \cellcolor{yellow!50}$0.40^{+0.74}_{-0.74}$ & \cellcolor{yellow!50}$0.62^{+1.1}_{-1.1}$ & -- & -- & -- & -- & -- & -- & -- & -- & -- & -- & -- \\
Engagement Depth Lift (1P) (\%) & -- & \cellcolor{green!50}$3.2^{+2.7}_{-2.7}$ & \cellcolor{green!50}$2.6^{+2.5}_{-2.4}$ & \cellcolor{yellow!50}$2.1^{+3.9}_{-3.8}$ & -- & -- & -- & -- & -- & -- & -- & -- & -- & -- & -- \\
Engagement Breadth Lift (AI Studio) (\%) & -- & -- & -- & -- & -- & -- & -- & \cellcolor{yellow!50}$-0.38^{+1.33}_{-1.32}$ & \cellcolor{green!50}$1.73^{+1.03}_{-1.02}$ & \cellcolor{green!50}$2.09^{+1.16}_{-1.15}$ & \cellcolor{green!50}$4.47^{+1.01}_{-1.00}$ & \cellcolor{yellow!50}$0.05^{+0.43}_{-0.45}$ & \cellcolor{green!50}$2.12^{+0.56}_{-0.56}$ & \cellcolor{green!50}$8.8^{+2.4}_{-2.3}$ & \cellcolor{yellow!50}$0.41^{+0.53}_{-0.53}$ \\
Engagement Depth Lift (AI Studio) (\%) & -- & -- & -- & -- & -- & -- & -- & \cellcolor{yellow!50}$3.52^{+4.47}_{-4.29}$ & \cellcolor{yellow!50}$1.38^{+4.62}_{-4.41}$ & \cellcolor{green!50}$3.8^{+3.5}_{-3.4}$ & \cellcolor{green!50}$18.2^{+2.5}_{-2.4}$ & \cellcolor{red!50}$-2.9^{+2.7}_{-2.7}$ & \cellcolor{green!50}$19.4^{+2.3}_{-2.2}$ & \cellcolor{green!50}$11.2^{+4.0}_{-3.8}$ & \cellcolor{green!50}$5.0^{+1.0}_{-1.0}$ \\
\midrule
\multicolumn{16}{l}{\textit{\textbf{Comparison Metrics}}} \\
Human Win Rate vs GPT-4o (\%) & -- & 38.5 & 37.4 & 39.6 & 44.1 & 44.5 & 46.2 & -- & -- & -- & -- & -- & -- & -- & -- \\
Human Win Rate vs Prev. Model (\%) & -- & 51.3 & 50.4 & 50.2 & 52.0 & 51.3 & 52.5 & -- & -- & -- & -- & -- & -- & -- & -- \\
Pointwise RM Win Rate (Internal) (\%) & -- & 53.6 & 56.5 & 57.4 & 57.6 & 56.2 & 56.5 & 49.7 & 59.4 & 65.8 & 63.7 & 43.7 & 57.3 & 57.3 & 50.3 \\
Pointwise RM Win Rate (User) (\%) & -- & -- & -- & -- & -- & -- & -- & 52.2 & 53.6 & 63.7 & 66.5 & 70.7 & 58.8 & 68.1 & 55.7 \\
Pairwise RM Win Rate (User) (\%) & -- & -- & -- & -- & -- & -- & -- & -- & -- & -- & -- & 66.8 & 59.9 & 64.0 & 49.1 \\
\midrule
\multicolumn{16}{l}{\textit{\textbf{Response Characteristics}}} \\
False Refusal (Internal) (\%) & 19.6 & 20.4 & 18.9 & 17.0 & 29.0 & 23.6 & 22.0 & 21.5 & 19.8 & 16.8 & 17.7 & 15.5 & 19.1 & 13.0 & 14.6 \\
False Refusal (User) (\%) & -- & -- & -- & -- & -- & -- & -- & -- & 48.8 & 30.7 & 28.5 & 26.5 & 19.3 & 14.9 & 20.4 \\
Avg Response Length (tokens) & 51.9 & 53.2 & 52.3 & 58.6 & 58.7 & 59.7 & 58.9 & 67.8 & 55.2 & 45.2 & 50.8 & 50.2 & 42.7 & 40.6 & 44.8 \\
Contains List (\%) & -- & 3.9 & 4.5 & 6.3 & 7.1 & 4.3 & 4.5 & 1.2 & 10.0 & 6.0 & 7.6 & 9.9 & -- & 13.2 & 13.2 \\
Contains Emoji (\%) & 27.8 & 25.1 & 24.7 & 22.5 & 20.5 & 26.7 & 26.1 & 22.7 & 10.3 & 20.2 & 15.9 & 16.7 & -- & -- & -- \\
Preachy Tone (\%) & -- & -- & 26.9 & 23.4 & 15.9 & 18.6 & -- & -- & -- & -- & -- & -- & -- & -- & -- \\
Positive Sentiment (\%) & 47.3 & 47.3 & 47.4 & 49.7 & 54.1 & 55.1 & 55.6 & 63.0 & -- & -- & -- & -- & -- & -- & -- \\
Instruction Violation (\%) & -- & 26.6 & 22.6 & 17.9 & 22.2 & 13.7 & 7.5 & 5.8 & -- & -- & -- & -- & -- & -- & -- \\
Cooperative Ratio (\%) & -- & -- & 75.1 & 80.6 & 78.7 & 90.8 & 92.3 & 91.0 & 89.9 & 90.6 & 90.5 & 87.9 & 87.9 & -- & -- \\
Non-Preachy Rate (\%) & -- & -- & -- & 76.3 & 82.3 & 86.7 & 90.9 & 88.5 & 92.4 & 95.0 & 94.9 & -- & -- & -- & -- \\
Templated Responses (\%) & -- & -- & -- & 10.3 & 7.9 & -- & -- & -- & -- & -- & -- & -- & -- & -- & -- \\
Wall-of-Text Failure (\%) & -- & -- & -- & -- & 12.2 & 9.3 & 9.4 & 8.5 & 5.6 & 2.0 & 2.4 & 1.3 & 6.8 & 0.9 & -- \\
\midrule
\multicolumn{16}{l}{\textit{\textbf{Language Match Rates}}} \\
Hindi (India) & -- & -- & -- & -- & -- & -- & -- & -- & -- & -- & -- & -- & 0.621 & 0.833 & 0.856 \\
Romanized Hindi (India) & -- & -- & -- & -- & -- & -- & -- & -- & -- & -- & -- & -- & 0.448 & 0.902 & 0.866 \\
Spanish & -- & -- & -- & -- & -- & -- & -- & -- & -- & -- & -- & -- & 0.582 & 0.919 & 0.980 \\
\midrule
\multicolumn{16}{l}{\textit{\textbf{Technical Benchmarks}}} \\
MATH (\%) & -- & 35.4 & 42.3 & 39.4 & 45.1 & 47.1 & 50.5 & 55.4 & 55.8 & 57.1 & 58.3 & 58.3 & 57.3 & 58.7 & -- \\
GSM8K (\%) & -- & 80.0 & 86.7 & 85.6 & 89.6 & 90.7 & 92.3 & 93.2 & 92.8 & 92.7 & 93.7 & 93.3 & 93.3 & 93.6 & -- \\
MBPP (\%) & -- & 65.2 & 64.6 & 64.8 & 54.8 & 65.2 & 66.6 & 70.0 & 69.8 & 69.0 & 66.6 & 59.2 & 62.6 & 55.8 & -- \\
HumanEval (\%) & -- & 63.4 & 64.6 & 66.5 & 75.0 & 75.0 & 77.4 & 76.2 & 78.7 & 74.4 & 73.8 & 75.6 & 79.3 & 76.2 & -- \\
HellaSwag (\%) & -- & 86.3 & 89.1 & 89.7 & 90.9 & 91.2 & 91.1 & 90.7 & 90.4 & 90.9 & 91.0 & 90.8 & 90.9 & 90.5 & -- \\
MMLU (\%) & -- & 72.3 & 80.3 & 78.6 & 80.1 & 80.1 & 79.5 & 83.2 & 83.1 & 84.1 & 85.2 & 85.1 & 84.9 & 84.5 & -- \\
ARC Challenge (\%) & -- & 85.7 & 93.4 & 93.2 & 93.4 & 93.7 & 93.1 & 95.3 & 95.4 & 95.1 & 94.9 & 94.8 & 95.2 & 95.4 & -- \\
ARC Easy (\%) & -- & 95.5 & 98.1 & 98.0 & 97.8 & 98.1 & 98.0 & 98.5 & 98.4 & 98.2 & 98.1 & 98.1 & 97.8 & 98.0 & -- \\
GPQA (\%) & -- & 26.5 & 41.3 & 39.3 & 42.4 & 42.2 & 39.3 & 42.0 & 42.0 & 46.0 & 44.6 & 43.8 & 44.4 & 44.2 & -- \\
IFEval (\%) & -- & 59.2 & 68.6 & 70.3 & 74.4 & 84.8 & 84.8 & 81.1 & 83.1 & 83.2 & 85.9 & 83.7 & 85.5 & 86.7 & -- \\
\bottomrule
\end{tabular}
}
\end{table*}

\section{Confidence Intervals for Ratio Metrics}
\label{appendix:ci}
Let $\hat{\mu}_T$ and $\hat{\mu}_C$ denote the empirical means for the test and control groups, respectively. We assume these estimators are independent (due to randomized assignment) and follow an approximate normal distribution. Let $\sigma_T^2$ and $\sigma_C^2$ denote the estimated variances of these means (the squared standard errors):
\begin{align}
    \sigma_g^2 = \widehat{\text{Var}}(\hat{\mu}_g) = \frac{S_g^2}{n_g},
\end{align}
where $S_g^2$ is the sample variance of the user-level outcomes in group $g$. Note that by calculating the variance of user-level aggregates (e.g., the user's average daily participation $\bar{Y}_i$ or total sends $S_i$), our estimator implicitly accounts for any within-user temporal autocorrelation over the readout window.

We seek the $100(1-\alpha)\%$ confidence interval for the true ratio $\rho = \mu_T / \mu_C$. Based on Fieller's theorem, the set of plausible ratios $r$ satisfies the condition:
\begin{align}
    \frac{(\hat{\mu}_T - r \hat{\mu}_C)^2}{\sigma_T^2 + r^2 \sigma_C^2} \le z^2,
\end{align}
where $z = z_{1-\alpha/2}$ is the standard normal critical value (e.g., $z \approx 1.96$ for 95\% confidence). Solving this quadratic inequality for $r$ yields the closed-form bounds:
\begin{align}
    R_{\pm} = \frac{\hat{\mu}_T \hat{\mu}_C \pm z \sqrt{\hat{\mu}_T^2 \sigma_C^2 + \hat{\mu}_C^2 \sigma_T^2 - z^2 \sigma_T^2 \sigma_C^2}}{\hat{\mu}_C^2 - z^2 \sigma_C^2}.
\end{align}
The confidence interval for the percentage lift is then reported as:
\begin{align}
    \text{CI}_{\text{Lift}}(\%) = \Big[ 100(R_{-} - 1), \; 100(R_{+} - 1) \Big].
\end{align}
Statistical significance at level $\alpha$ is declared if the null lift (0\%) is not contained within this interval.

\end{document}